\documentclass[]{youtu} % For LaTeX2e

\usepackage{mathpazo}
\usepackage{graphicx}
\usepackage[numbers]{natbib}
\usepackage{algorithm}
\usepackage{algorithmic}
\usepackage[table,dvipsnames]{xcolor}
\tcbuselibrary{listings, breakable}
\usepackage{booktabs,multirow,subcaption}
\usepackage{multirow}
\usepackage{amssymb}
\usepackage{makecell}

\usepackage{hyperref}
\usepackage{url}
\usepackage{tcolorbox}
\tcbuselibrary{breakable}
\usepackage{subcaption}
\usepackage{adjustbox}

\usepackage{tabularx}
\usepackage{booktabs}
\usepackage{multirow}
\usepackage{array}
\usepackage{xcolor}

\lstdefinestyle{pythonstyle}{
    language=Python,
    basicstyle=\ttfamily\small,
    numbers=none,
    numberstyle=\tiny\color{gray},
    keywordstyle=\color{blue!70!black},
    commentstyle=\color{green!50!black},
    stringstyle=\color{orange!80!black},
    breaklines=true,
    tabsize=2,
    showstringspaces=false,
}

\newtcblisting{pycodebox}{
  listing engine=listings,
  colback=blue!5!white,             % 淡蓝背景
  colframe=MidnightBlue!85!black,  % 深蓝边框
  coltitle=white,                  
  colbacktitle=MidnightBlue!85!black,
  title=\textbf{Python Code},
  fonttitle=\bfseries,
  listing only,
  breakable,
  listing options={style=pythonstyle},
}

\newtcolorbox{outputbox}{
  colback=gray!10!white,       % 浅灰背景
  colframe=MidnightBlue!80!black, % 深蓝边框
  fonttitle=\bfseries,
  coltitle=white,
  colbacktitle=MidnightBlue!80!black,
  title=Output,
  left=4mm, right=4mm, top=2mm, bottom=2mm,
  boxrule=0.8pt,
  arc=3mm, % 圆角
}

\newtcolorbox{databox}[2][]{
  enhanced,
  breakable,
  colback=blue!3!white,
  colframe=MidnightBlue!75!black,
  coltitle=white,
  colbacktitle=MidnightBlue!75!black,
  fonttitle=\bfseries,
  title={#2},
  boxrule=0.8pt,
  arc=2.5mm,
  left=2mm,right=2mm,top=1.5mm,bottom=1.5mm,
  #1
}

\newtcolorbox{fieldbox}[2][]{
  enhanced,
  breakable,
  colback=white,
  colframe=MidnightBlue!40,
  boxrule=0.5pt,
  arc=1.5mm,
  fonttitle=\bfseries,
  title={#2},
  left=1.5mm,right=1.5mm,top=1mm,bottom=1mm,
  #1
}

\newtcolorbox{subfieldbox}[2][]{
  enhanced,
  breakable,
  colback=gray!6!white,
  colframe=black!20,
  boxrule=0.5pt,
  arc=1mm,
  fonttitle=\bfseries,
  title={#2},
  left=1mm,right=1mm,top=0.8mm,bottom=0.8mm,
  #1
}

\definecolor{hlblue}{RGB}{182,210,232} % 近似图中的浅蓝

\title{Internalizing the Future: A Unified Agentic Training Paradigm for World Model Planning}

\author{Xuan Zhang$^{1,2,3}$, Zhijian Zhou$^{1,2,3}$, Lingfeng Qiao$^{3}$, Yulei Qin$^{3}$, Ke Li$^{3}$, Xing Sun$^{3}$, Xiaoyu Tan$^{3\dagger}$, Chao Qu$^{1\dagger}$, Yuan Qi$^{1\dagger}$} 

\affiliation{$^{1}$Fudan University, $^{2}$Shanghai Innovation Institute, $^{3}$Tencent Youtu Lab}

\date{June 25, 2026}
\correspondence{xuanzhang24@m.fudan.edu.cn}
\begin{document}

\abstract{Large language model (LLM) agents have demonstrated strong capability in sequential decision-making, yet they remains fundamentally reactive in long-horizon tasks. Unlike humans who employ \textit{"what-if"} reasoning to evaluate potential plans before commitment, standard agents lack an internal world model to simulate future outcomes. Therefore, we propose to internalize future-aware planning by training a single autoregressive model to verbalize both a prospective state rollout and a plan-conditioned success estimate—a textual analogue of the $Q$-value. Crucially, we identify a format-capability gap: simply fine-tuning agents on look-ahead traces during post-training leads to superficial mimicry of foresight without genuine predictive grounding. To bridge this gap, we introduce a three-stage training paradigm: (i) World Model Agentic Mid-Training (WM-AMT) to inject latent predictive capabilities into the policy; (ii) Format-Eliciting SFT (FE-SFT) to structure this injected capability; and (iii) Foresight-Conditioned Reinforcement Learning (FC-RL) to refine the calibration and utility of the generated simulations. Evaluated on search and mathematical reasoning tasks, our approach consistently outperforms other training baselines. Our results demonstrate that effective internal world modeling in LLM agents requires a capability-first training pipeline to achieve grounded and calibrated foresight.}

\maketitle

% \dateandcorrespondence

\section{Introduction}
\label{sec:introduction}
\begin{figure*}
  \centering
  \includegraphics[width=0.8\linewidth]{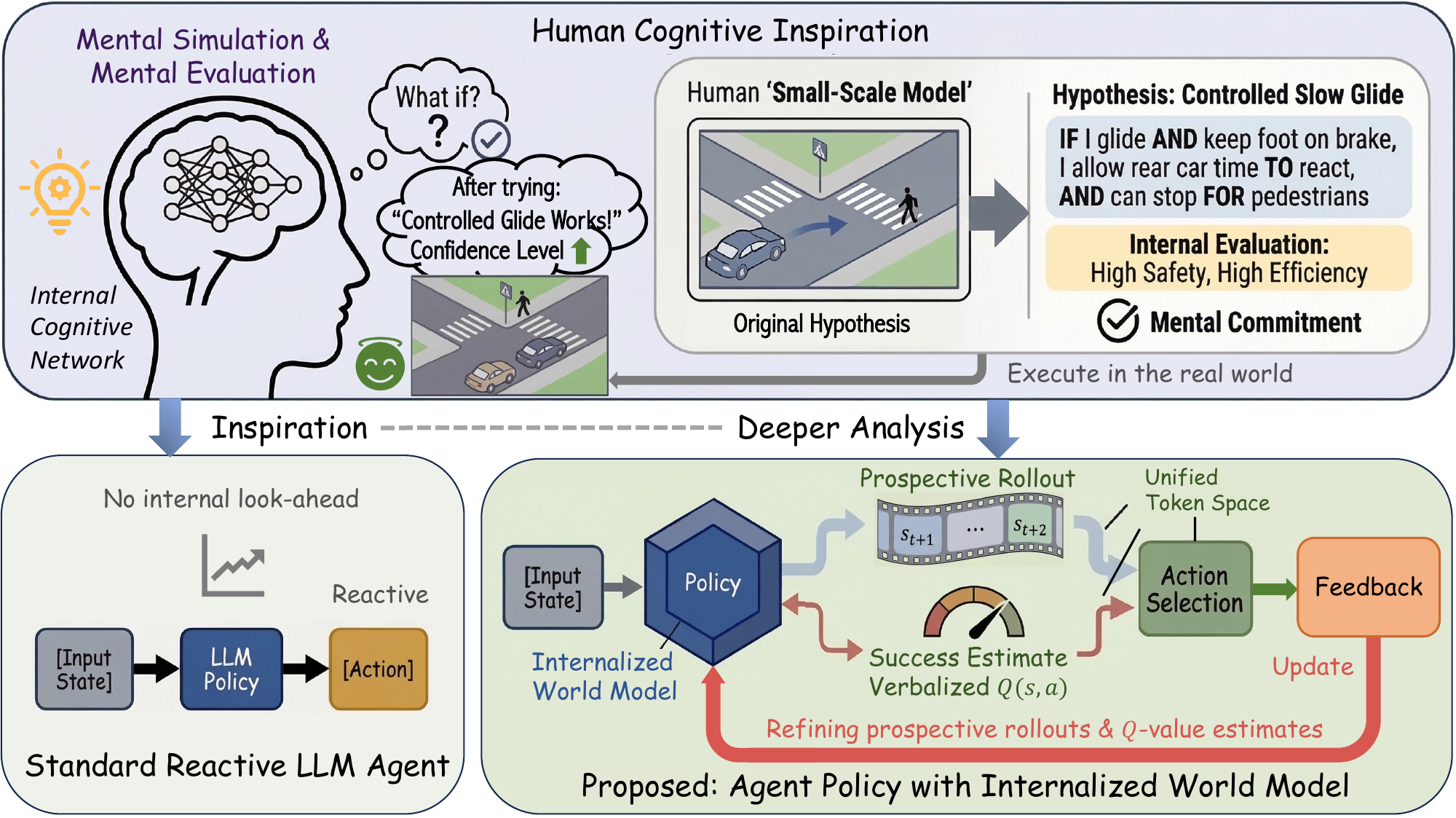}
  \caption{Overview of our key design of an agent policy with internalized world model.}
  \label{intro}
\end{figure*}

LLM agents have become strong controllers for multi-step tasks given their comprehensive pre-trained semantic knowledge and remarkable capabilities in sequential decision-making \citep{Wang2023ASO, sumers2024cognitive}. However, the prevailing paradigm in long-horizon settings remains fundamentally reactive, yet human planning is rarely reactive. Before acting, people often ask a \textit{what-if} question: if I follow this plan, what is likely to happen next, and how likely is it to succeed? This style of reasoning is central to intelligent behavior because it allows alternatives to be mentally compared before commitment, as shown in Figure \ref{intro}. In Craik’s classic formulation, effective intelligence depends on carrying a “small-scale model” of external reality that lets possible actions be tried out in thought before they are executed \citep{Craik1944TheNO}. Inspired by this pattern of planning, we do not claim full counterfactual reasoning; rather, we target a plan-conditioned form of future simulation and evaluation within a language agent.

While standard autoregressive reasoning methods such as Chain-of-Thought \citep{NEURIPS2022_9d560961} mainly improve decomposition of the current problem, the ability that we aim to inject is closely related to an internal world model \citep{10.1145/122344.122377, ha2018recurrent}. Instead of utilizing separate modules to serve as external world models \citep{chae2025web, wang2025llmsscalablegeneralpurposesimulators, Gu2025WebDreamer}, our key claim is that, for language agents, this kind of future-aware planning signal should be internalized as part of the policy itself. Moreover, unlike Dreamer-style model-based RL \citep{Hafner2020Dream, hafner2021mastering, hafner2024masteringdiversedomainsworld}, our goal is not to introduce a separate simulator or a dedicated value head. Instead, we train a single autoregressive model not only to serve as a reasoning agent, but also to verbalize both: (i) a compact prospective rollout of what may happen next, and (ii) a plan-conditioned estimate of success if the current plan is followed. This textual estimate is best understood as a verbalized analogue of $Q(s, a)$ or local action value, represented in the same token space as the policy rather than as a claim of full what-if reasoning. Unlike other methods that only train agents to predict future states or transition dynamics \citep{chen2025internalizingworldmodelsselfplay, cwm2025, zhang2025agentlearningearlyexperience}, our design of internal world modeling is more comprehensive and innovative. Crucially, our verbalized $Q$-values are designed to be continuously updated and corrected by actual environmental feedback obtained from real execution during RL training, which is also distinct from previous works \citep{hao-etal-2023-reasoning}.

Such explicit verbalization offers a transparent reasoning signal, allowing the agent to effectively prune suboptimal paths through implicit deliberation. To inject this capability, a natural idea is to leverage the inherent world modeling potential of LLMs \citep{li2025wordworldlargelanguage} by training the agent to imagine the future, evaluate it, and then act. However, our preliminary experiments show that this strategy fails when applied only during post-training. With SFT and RL on formatted look-ahead traces, the agent learns the surface form of a simulation but not the predictive competence behind it. Experimental results for SFT and RL are shown in Table \ref{tab:preliminary_results}, and an example is shown in Appendix \ref{subsec: gap_example}. We call this failure mode the \textit{format-capability gap}: the agent can produce a plausible-looking foresight block while the content remains weakly grounded, miscalibrated, or outright misleading, thus introducing noise rather than useful guidance. 

This motivates us to propose a unified three-stage training paradigm: \textbf{World Model Agentic Mid-Training (WM-AMT)}, \textbf{Format-Eliciting SFT (FE-SFT)}, and \textbf{Foresight-Conditioned Reinforcement Learning (FC-RL)}. WM-AMT injects the underlying predict-then-act capability by augmenting large-scale agentic trajectories with future-aware experience summaries. FE-SFT then teaches the model when and how to externalize this latent capability in a structured foresight block. FC-RL finally improves both action quality and foresight quality by rewarding task success, grounded anticipation, and calibrated confidence. All three stages are implemented within one causal language model, with no additional value head.

We evaluate this training paradigm on two representative agentic domains: search and mathematical reasoning. Across both settings, our method consistently outperforms other training baselines. The results show that future-aware behavior in LLM agents is not well served by format supervision alone; it benefits from a training pipeline that first builds the capability, then elicits it, and finally optimizes it under reward.

Our contributions are summarized as follows:

\begin{itemize}
    \item We identify the format-capability gap: post-training on look-ahead formats alone teaches LLM agents to mimic the structure of foresight without reliably acquiring grounded future prediction.
    \item We propose a unified three-stage paradigm, consisting of WM-AMT, FE-SFT, and FC-RL, which internalizes world-model planning inside a single autoregressive policy. Our method verbalizes both future rollout and a plan-conditioned success estimate without introducing an explicit value head, and this verbalized Q-like signal is explicitly trained and refined throughout the pipeline.
    \item We show on search and mathematical reasoning that this capability-first pipeline yields consistent gains over other training baselines.
\end{itemize}

\section{Preliminaries}
\subsection{World models in LLM agents}

We formulate LLM agents' decision-making as a Markov Decision Process (MDP) \citep{Bellman}, denoted by $\mathcal{M} = \langle \mathcal{S}, \mathcal{A}, \mathcal{P}, \mathcal{R}, \gamma \rangle$. State $s_t \in \mathcal{S}$ includes the initial instruction, historical trajectory, and observations. Action $a_t \in \mathcal{A}$ represents discrete operations, e.g., text generation or tool invocation. Transition dynamics $\mathcal{P}(s_{t+1}|s_t, a_t)$ and reward $\mathcal{R}(s_t, a_t)$ govern environmental feedback. The objective is to learn a parameterized policy $\pi_\theta(a_t|s_t)$ that maximizes the expected return discounted by $\gamma \in [0, 1]$. World models traditionally predict transition dynamics $\hat{\mathcal{P}}(s_{t+1}|s_t, a_t)$ and rewards $\hat{\mathcal{R}}(s_t, a_t)$ \citep{10.1145/122344.122377}. However, applying token-level simulation to dense, low-level environmental feedback is computationally intractable. Autoregressively predicting fine-grained future trajectories $\tau_{future} = \langle s_{t+1}, a_{t+1}, o_{t+1}, \dots, s_{t+k} \rangle$ may waste computational capacity on syntax rather than core causal dynamics, thus causing severe compounding errors \citep{lecun2022path}. Thus, we reconceptualize the World Model for LLM agents as an \textit{internal semantic simulator}. It generates a condensed \textit{experience summary} of anticipated consequences, defined as $\hat{z}_t = \Psi(\tau_{future})$. Here, $\Psi$ is a semantic compression function mapping raw rollouts into high-level predictive foresight. This empowers the agent to perform \textit{look-ahead planning}, transitioning from a reactive policy $\pi_\theta(a_t|s_t)$ to a foresight-conditioned policy $\pi_\theta(a_t|s_t, \hat{z}_t)$. Ultimately, this enables the agent to evaluate consequences and prune suboptimal trajectories in the latent space before executing real-world actions.

\subsection{Mid-training in large language models}

To obtain strong initial policies in terms of both performance and exploration diversity, mid-training \citep{mo2025midtraininglargelanguagemodels, su2025scalingagentscontinualpretraining} has become an essential step between general pre-training and post-training, which aims to adapt the LLM's underlying data distribution toward the target domain. Recent theoretical analyses demonstrate that an effective mid-training stage not only establishes strong policy priors by efficiently pruning the decision space, but also allows the subsequent RL phase to effectively jump across multiple steps during Bellman backups, thus accelerating convergence speed and minimizing post-training regret \citep{zhang2026learning}. By immersing the LLM in high-density, high-quality specialized sequences, mid-training provides a latent semantic foundation and contextually grounded initialization for post-training.

\subsection{Group-based RL}

 Unlike traditional Actor-Critic methods (e.g., PPO \citep{schulman2017proximalpolicyoptimizationalgorithms}) that rely on a parameterized value network $V_\phi(s)$ to estimate advantages, group-based RL approaches \citep{deepseek-math, yu2025dapo, feng2025groupingroup} leverage the statistical distribution of rewards within a sampled batch to approximate the baseline. For a given task prompt $x$, the agent samples a group of $G$ trajectories $\{\tau_1, \tau_2, \dots, \tau_G\}$ from the current policy $\pi_{\theta_{\text{old}}}$. Each trajectory $\tau_i$ represents a full episode of interactions and is assigned a scalar reward $R(\tau_i)$ reflecting the task success. The advantage estimation $A(\tau_i)$ is derived from the normalized reward in each group: $A(\tau_i) = \frac{R(\tau_i) - \mu(\{R(\tau_j)\}_{j=1}^G)}{\sigma(\{R(\tau_j)\}_{j=1}^G)}$, where $\mu$ and $\sigma$ denote the mean and standard deviation of rewards in the group, respectively.

\section{Methodology}
\label{sec:methodology}

\subsection{The format-capability gap in post-training}
\label{subsec: gap}
\begin{figure*}
  \centering
  \includegraphics[width=1.0\linewidth]{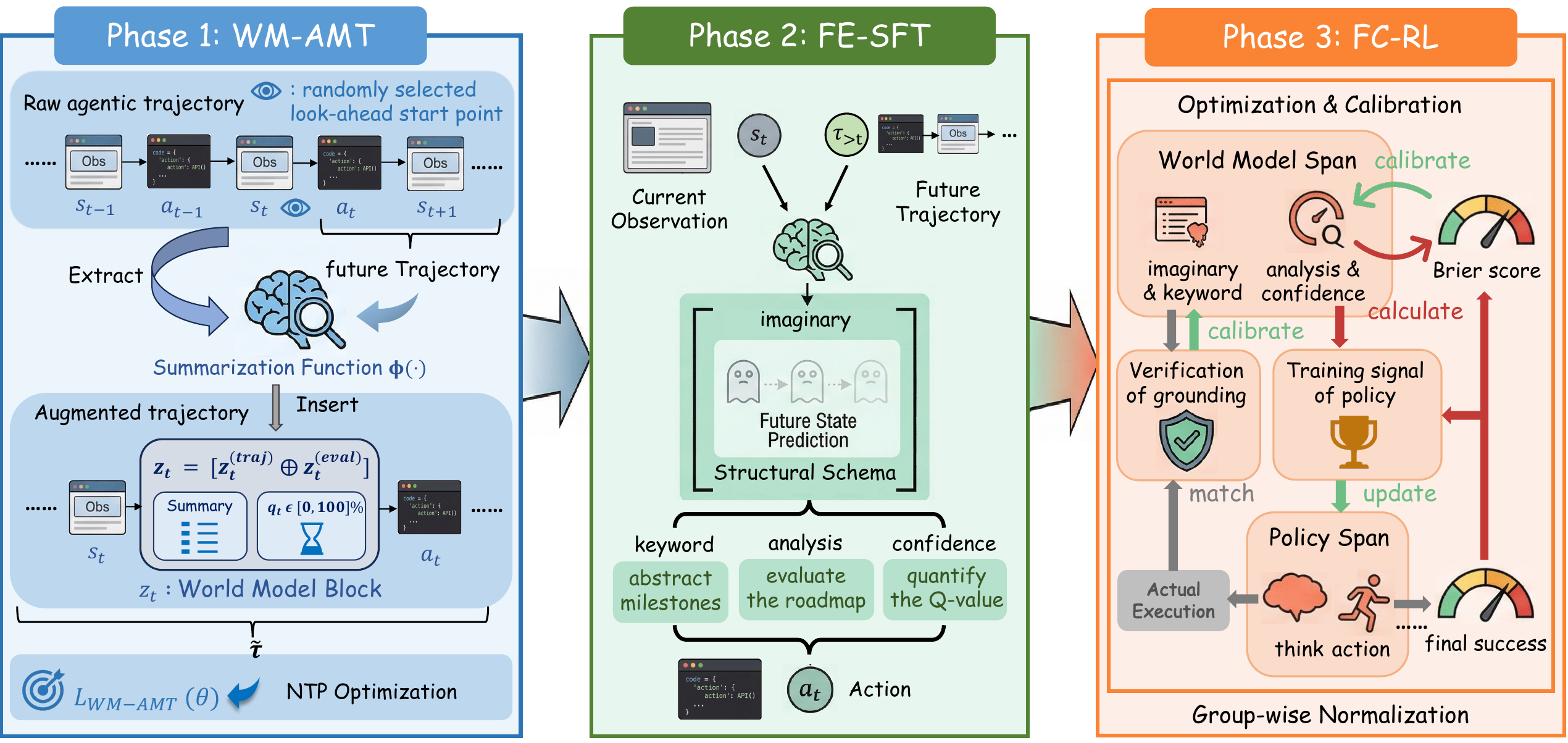}
  \caption{Overview of our three-stage training paradigm.}
  \label{fig1}
\end{figure*}

A natural idea to internalize world model into the LLM policy is training the LLM with SFT data that explicitly includes anticipated futures, and then followed by RL. However, our preliminary experiments on the model mid-trained with routine agentic trajectories (see Appendix \ref{appendix: gap_results} and \ref{subsec: gap_example}) reveal a format-capability gap. While the agent learns the format of look-ahead reasoning, it fails to simulate accurate future dynamics, often resulting in hallucinations. We attribute it to the inherent nature of post-training: while SFT and RL excel at \textit{eliciting} latent capabilities, they struggle to \textit{inject} fundamentally new ones. Without a prior understanding of future consequences, enforcing a look-ahead format merely triggers hallucinated pattern matching rather than genuine future simulation. It motivates us to intervene earlier in the training pipeline via \textbf{mid-training} to construct a foundational data distribution that intrinsically models the future. We propose a three-stage training paradigm to close the format-capability gap, as shown in Figure \ref{fig1}. Given an LLM policy, our goal is not only to produce a well-formed look-ahead trace, but also to internalize transition-aware foresight that improves policy capability. The paradigm consists of: (i) \textbf{World Model Agentic Mid-Training (WM-AMT)} for policy prior formation, (ii) \textbf{Format-Eliciting Supervised Fine-Tuning (FE-SFT)} for structured elicitation, and (iii) \textbf{Foresight-Conditioned Reinforcement Learning (FC-RL)} for reward-aligned policy optimization.

% A natural idea to internalize world model into the LLM policy is to train it with SFT data where the agent explicitly generates the anticipated future before outputting the next action, and then followed by RL. However, integrating the \textit{look-ahead} capability into LLM agents via this standard post-training paradigms presents significant challenges. As shown in our preliminary experimental results in Appendix \ref{subsec: gap_example}, directly applying SFT and RL on the routinely mid-trained model with such reasoning-enhanced data leads to suboptimal performance. While the model successfully learns the syntactic format, it fails to acquire the underlying semantic capability to accurately simulate future dynamics. The formatted future predictions often hallucinates or diverges from the ground truth, effectively introducing noise rather than useful guidance into the reasoning process. We term this failure mode as the \textbf{format-capability gap}, and attribute it to the inherent nature of post-training: while SFT and RL excel at \textit{eliciting} latent capabilities, they struggle to \textit{inject} fundamentally new ones. Without a prior understanding of future consequences, enforcing a look-ahead format merely triggers hallucinated pattern matching rather than genuine future simulation. 

\subsection{World model agentic mid-training}
\label{subsec:mid_training}

Standard pre-training and agentic SFT trajectory data primarily consist of observational trajectories. While this paradigm trains the model to predict the next action given the historical context, we aim to instill look-ahead capabilities in LLM agents via WM-AMT (Algorithm \ref{alg:wm_amt}). Let $\mathcal{D}_{\text{raw}} = \{ \tau_i \}_{i=1}^N$ denote a large-scale dataset of agentic trajectories, where each trajectory is defined as a sequence of state-action pairs: $\tau = (s_1, a_1, s_2, a_2, \dots, s_T, a_T)$. Instead of relying on human-annotated chain-of-thought data, we systematically augment $\mathcal{D}_{\text{raw}}$ through a trajectory summarization mapping.

Formally, for a given trajectory $\tau$ of length $T$, we sample a set of indices $\mathcal{I} \sim \mathcal{U}(1, T-1)$. For each step $t \in \mathcal{I}$, we construct a structured world model block $z_t$. We define a summarization function $\Phi(\cdot)$, which maps the current observation $s_t$ and the future trajectory $\tau_{>t}$ into a composite textual representation: $z_t = [z_t^{(\text{traj})} \oplus z_t^{(\text{eval})}]$. Here, $z_t^{(\text{traj})}$ serves as an abstracted roadmap of the simulated future. $z_t^{(\text{eval})}$ provides a semantic gap analysis and a $Q$-value estimation $q \in [0, 100]\%$, denoting the confidence score of successfully resolving the initial task following this roadmap. While accurately estimating $q$ for merely a single step of a trajectory ideally requires dozens of Monte Carlo rollouts, it is computationally prohibitive given the massive scale of mid-training data. Consequently, we adopt an LLM to heuristically estimate this success probability. Though this estimation may be not accurate enough, it is acceptable because these confidence priors are ultimately grounded and calibrated against true execution outcomes during subsequent RL training.

We then inject $z_t$ into the original agentic trajectory, producing an augmented trajectory $\tilde{\tau}$, which is defined by $\tilde{\tau} = (\dots, s_t, z_t, a_t, s_{t+1}, \dots) $. The WM-AMT stage optimizes the policy $\theta$ via the standard autoregressive negative log-likelihood objective over the augmented dataset $\mathcal{D}_{\text{WM-AMT}}$:
$
\mathcal{L}_{\text{WM-AMT}}(\theta) = - \mathbb{E}_{\tilde{\tau} \sim \mathcal{D}_{\text{WM-AMT}}} \left[ \sum_{i=1}^{|\tilde{\tau}|} \log P_\theta(u_i | u_{<i}) \right]
$, 
where $u_i$ denotes the $i$-th token in $\tilde{\tau}$. By forcing the model to autoregressively reconstruct the composite block $z_t$ conditioned strictly on the history up to $s_t$, the objective functions as an implicit world model, equipping the agent with both causal prediction and value evaluation capabilities.

\subsection{Format-eliciting supervised fine-tuning}
\label{subsec:sft}

While WM-AMT successfully injects latent look-ahead capabilities into LLM agents, the base policy lacks the structural schema to autonomously and reliably invoke this predictive capability during inference. The FE-SFT stage bridges this gap by aligning the generated trajectory with a predefined, explicit look-ahead reasoning structure. We curate a filtered task-specific instruction-tuning dataset $\mathcal{D}_{\text{FE-SFT}}$. Given a current observation $s_t$, the policy is trained to project its internal world model into a textual representation of anticipated future states and estimated $Q$-values, i.e., world model block $\hat{z}_t$. While adopting a similar generation pipeline for these instances, FE-SFT strictly encloses $\hat{z}_t$ within distinct structural delimiters. Following $\hat{z}_t$, the policy generates the corresponding executable action $a_t$. Since WM-AMT has established a robust causal prior via a large-scale general agentic trajectories, FE-SFT avoids the hallucinations typical of world modeling from scratch. We also use the standard autoregressive negative log-likelihood objective to optimize the agent on $\mathcal{D}_{\text{FE-SFT}}$.

\subsection{Foresight-conditioned RL}
\label{subsec:rl}

The policy may still suffer from compounding errors during long-horizon inference or produce miscalibrated foresight after WM-AMT and FE-SFT. 
To jointly optimize the internalized world model and the policy, we introduce FC-RL (Algorithm \ref{alg:fg_rl}). The agent serves as two roles and generates a structured trajectory in an interaction loop: 
$\underbrace{\textit{imaginary} \rightarrow \textit{keyword} \rightarrow \textit{analysis} \rightarrow \textit{confidence} \ (q_t)}_{\text{World Model Span } (\mathcal{T}_{\text{WM}})} \rightarrow \underbrace{\textit{think} \rightarrow \textit{action} \ (a_t)}_{\text{Policy Span } (\mathcal{T}_{\text{Policy}})}$.

To jointly optimize final task success and intermediate foresight quality, we decompose the training objective into world model optimization and policy optimization. For a given prompt, we sample a group of $N$ trajectories $\{ \tau_1, \dots, \tau_N \}$.

\paragraph{World model optimization: foresight grounding and calibration}
To ensure the internal world model provides reliable planning rather than hallucinations, we optimize it from two complementary dimensions: causal accuracy and value calibration.

First, to penalize hallucinated simulations, we introduce a grounding reward $R_{\text{ground}}$. The agent abstracts its look-ahead trajectory into milestones, denoted as keywords $K_t$. We define $R_{\text{ground}}^{i,t} = R_{\text{match}}(K_t^{(i)}, \tau_{>t}^{(i)})$, a verification function that rewards the model only if its predicted keywords successfully manifest in the actual execution.

Second, to ensure the accuracy of the agent's internal $Q$-value estimation and explicitly penalize miscalibrated foresight, we apply a Brier score penalty \citep{glenn1950verification} to the confidence $q_t^{(i)}$. Upon episode completion, the environment provides a terminal correctness evaluation, yielding a binary outcome indicator $\mathbf{I}$, i.e., $\mathbf{I}_{y \equiv y^*}$, where $y$ is the predicted answer and $y^*$ is the ground truth. From a Monte Carlo (MC) perspective, $\mathbf{I}$ serves as an unbiased, single-sample empirical return for the trajectory from step $t$. To calibrate $q_t$, we treat this MC return as the ground truth and apply a Brier score-based calibration reward:

\begin{equation}
    R_{\text{calib}}^{i, t}=R_{\text{calib}}(y_i, q_t^{(i)}, y_i^*) = \mathbf{I} - (q_t^{(i)} - \mathbf{I})^2
\end{equation}

By utilizing the terminal correctness to calibrate the intermediate $q_t$, it forces the agent to align its self-assessed $Q$-values, i.e., confidence on future, with the empirical success rate. 

Finally, we compute a combined step-level world model reward $R^{\text{WM}}(\tau_i^t) = R_{\text{calib}}^{i, t} + R_{\text{ground}}^{i,t}$. We obtain the final world model advantage $A(\tau[\text{world\_model}]_i^t)$ by normalizing these combined rewards across all the corresponding steps $\{R^{\text{WM}}(\tau_i^t)\}_{i=1}^N$ within the group.

\paragraph{Policy optimization: execution success}
For long-horizon tasks, relying solely on the terminal global advantage may suffer from sparse and delayed credit assignment, so we introduce a local (step) reward $R^S(\tau_i^t)=q_t^{(i)}$. Let $A^E(\tau_i)$ denote the global advantage computed by normalizing the episode rewards $R^E_i$ across the group. Similarly, let $A^S(\tau_i^t)$ represent the step-level advantage, normalized across all the confidence of corresponding steps $\{q_t^{(i)}\}_{i=1}^N$ within the group. The final combined advantage for the policy is defined as:

\begin{equation}
    A(\tau[\text{policy}]_i^t) = A^E(\tau_i) + \omega \cdot A^S(\tau_i^t)
\end{equation}

where $\omega$ balances global task success and local foresight alignment. Crucially, since $q_t^{(i)}$ is explicitly calibrated via world model optimization, it mitigates the risk of reward hacking and serves as a reliable proxy for the policy. The policy $\pi_\phi$ is then updated by maximizing the clipped GRPO surrogate objective utilizing the advantage $A(\tau[\text{policy}]_i^t)$.

\section{Related Work}
\subsection{World models in autonomous agents}
\label{ref: world_model}
World models are core to model-based RL, where agents learn transition and reward for look-ahead control \citep{10.1145/122344.122377, 10.1561/2200000086}. Classical lines focus on latent-dynamics learning in visually observed environments \citep{ha2018recurrent, Hafner2020Dream, hafner2021mastering, hafner2024masteringdiversedomainsworld} and more recently, embodied agent for future action-conditioned simulation \citep{fung2025embodiedaiagentsmodeling}. It enables models to improve sample efficiency in a latent space or simulate outcomes without real-world interaction. The idea has shifted from pixel-space simulation to language-mediated future prediction and planning due to the rise of LLMs \citep{hao-etal-2023-reasoning, chae2025web}; it also extends to code-generation settings, where world modeling is used to study long-horizon program synthesis and repair \citep{cwm2025}. Despite these advancements, most existing systems treat the world model as an explicit simulator or external module \citep{wang2025llmsscalablegeneralpurposesimulators, Gu2025WebDreamer}. For example, WebDreamer \citep{Gu2025WebDreamer} and WebEvolver \citep{fang2025webevolverenhancingwebagent} construct explicit world models to simulate the environmental feedback, which improve deliberation quality but may introduce additional inference overhead and possible simulator-policy mismatch. In parallel, recent implicit world modeling strategies suggest that early experiences can improve downstream agent behavior without relying on another simulator \citep{chen2025internalizingworldmodelsselfplay, zhang2025agentlearningearlyexperience}. Extending this implicit world modeling line, our work targets a unified agentic training pipeline for world model planning.

\subsection{Agentic data synthesis and mid-training}
The scarcity of expert-level demonstrations has motivated the synthetic data for training LLM agents \citep{zhang2025agentlearningearlyexperience}. Early imitation-based distillation from stronger teachers \citep{pahuja2025explorerscalingexplorationdrivenweb, deng2023mindweb} aids initial alignment but suffers from limited diversity and weak on-policy coverage. Subsequent works, such as AgentSynth \citep{xie2025agentsynth} and SCA \citep{zhou2025selfchallenginglanguagemodelagents}, broaden task synthesis to expand the exploration space for agentic RL. Another direction shifts from trajectory synthesis to environment synthesis, e.g., classical world-model methods in Section \ref{ref: world_model}, which uses learned environments to generate scalable interaction data. However, translating these synthetic interactions into reusable decision-making capabilities necessitates a dedicated training phase. Consequently, \textit{mid-training} \citep{mo2025midtraininglargelanguagemodels} is increasingly adopted to bridge general pre-training and task-level alignment \citep{su2025scalingagentscontinualpretraining, zhang2026learning}. Unlike SFT, mid-training scales to vastly larger, diverse corpora of expert rollouts, such as human demonstrations in device control, tool-use records and code writing \citep{rawles2023androidinthewild, huang-etal-2025-opencoder, NEURIPS2024_1704ddd0}. Our perspective is to synthesize agentic data with compressed future experience summary, and internalize a robust \textit{predict-then-act} policy prior via mid-training.

\section{Experiments}
\subsection{Experiment setup}
\label{subsec: setup}
We select an intermediate checkpoint of Youtu-LLM-2B \citep{lu2026youtullmunlockingnativeagentic} as our foundation model, and refer to it as Youtu-LLM-2B-Init. This selection is motivated by two practical considerations: (1) the Youtu-LLM family inherently possesses strong native agentic capabilities; (2) a lightweight 2B scale makes our extensive mid-training computationally feasible. To validate the effectiveness of our method, we initialize from the Youtu-LLM-2B-Init immediately following its general pre-training and compare two mid-training variants: (1) \textbf{Youtu-LLM-2B-Base} is trained with standard agentic mid-training on 200B tokens of high-quality agentic trajectory data without changing its structure; (2) \textbf{Youtu-LLM-2B-Base (WM-AMT, ours)} augments the same agentic trajectory data by inserting an explicit \textit{world model block} into a random split of multi-turn interaction trajectories, which provides supervised signals for future-state prediction. We conduct experiments on two representative agentic tasks: search and mathematical reasoning.

\paragraph{Search}
Following \citet{jin2025searchr}, we evaluate our training paradigm on 7 search-augmented QA tasks, including single-hop QA datasets (NQ \citep{47761}, TriviaQA \citep{joshi2017triviaqalargescaledistantly}, and PopQA \citep{mallen2023trustlanguagemodelsinvestigating}) and multi-hop QA datasets (HotpotQA \citep{yang2018hotpotqadatasetdiverseexplainable}, 2Wiki \citep{ho2020constructingmultihopqadataset}, MuSiQue \citep{trivedi2021musique}, and Bamboogle \citep{press2023measuringnarrowingcompositionalitygap}). For retrieval, we use the 2018 Wikipedia dump \citep{karpukhin-etal-2020-dense} as the knowledge source and E5 \citep{wang2024textembeddingsweaklysupervisedcontrastive} as the retriever, and set the number of retrieved passages to 3 across all retrieval-based methods. During evaluation, the results are obtained via greedy sampling, and the metrics are computed via LLM-as-a-judge using DeepSeek V3.1 \citep{deepseekai2025deepseekv3technicalreport}.

\paragraph{Mathematical reasoning}
Following \citet{feng2025retoolreinforcementlearningstrategic}, we evaluate our paradigm on AIME2024/2025/2026 \citep{aime}. We repeat the evaluation set AIME2024\&2025\&2026 30 times and report both mean@$30$ and pass@$30$. The evaluation hyperparameters of evaluation are set to temperature 1.0 and top-p 0.6.

\paragraph{Baselines and configurations}
(1) \textbf{Post-Training Only*:} Our preliminary experiments are conducted directly on Youtu-LLM-2B-Base, serving as a vanilla baseline that only applies post-training without any world-model-augmented mid-training. (2) \textbf{State-Only Prediction (Post-Training Only)$\dagger$:} The agent is trained only via post-training to predict future states but omits the verbalized $Q$-values. This setup mirrors other methods of predicting only state transitions \citep{chen2025internalizingworldmodelsselfplay}. (3) \textbf{State-Only Prediction (With Mid-Training)$\ddagger$:} The agent is trained via mid-training and post-training to predict future states but omits the verbalized $Q$-values. This setup mirrors other methods introducing large-scale mid-training for world modeling \citep{cwm2025}. (4) \textbf{IWM}: We apply the core world-modeling data generation pipeline of \citep{zhang2025agentlearningearlyexperience} to our task. Detailed training hyperparameters are presented in Appendix \ref{rl_config}.

\begin{table*}[t]
\centering % 确保整个双栏大表格在页面居中

% --- 子表 (a) --- 将 \textwidth 改为 0.9\textwidth
\begin{subtable}{0.9\textwidth}
\centering
% 将 width 同样改为 0.9\textwidth 
\begin{adjustbox}{width=0.9\textwidth,center}
\begin{tabular}{@{} l ccc cccc c @{}}
\toprule
\multirow{2}{*}{\textbf{Method}} & \multicolumn{3}{c}{\textbf{Single-Hop QA}} & \multicolumn{4}{c}{\textbf{Multi-Hop QA}} & \multirow{2}{*}{\textbf{Avg.}} \\
\cmidrule(lr){2-4} \cmidrule(lr){5-8}
& \textbf{NQ} & \textbf{TriviaQA} & \textbf{PopQA} & \textbf{HotpotQA} & \textbf{2Wiki} & \textbf{MuSiQue} & \textbf{Bamboogle} & \\
\midrule
\multicolumn{9}{c}{\textit{Youtu-LLM-2B-Base}} \\
\midrule
SFT* & 41.8 & 56.6 & 31.0 & 44.8 & 26.4 & 21.0 & 45.6 & 38.2 \\
FE-SFT* & 41.6 & 51.4 & 30.6 & 42.2 & 27.2 & 23.2 & 45.6 & 37.4 \\
IWM & 44.4 & 57.6 & 31.8 & \textbf{46.0} & 28.8 & \underline{25.4} & \textbf{47.2} & \underline{40.2} \\
\midrule
\multicolumn{9}{c}{\textit{Youtu-LLM-2B-Base (WM-AMT, ours)}} \\
\midrule
SFT & \textbf{46.0} & \underline{58.2} & \underline{32.6} & 44.6 & \underline{31.2} & 20.6 & \underline{46.4} & 39.9 \\
FE-SFT & \underline{45.6} & \textbf{58.8} & \textbf{33.8} & \underline{45.8} & \textbf{35.8} & \textbf{25.6} & \textbf{47.2} & \textbf{41.8} \\
\bottomrule
\end{tabular}
\end{adjustbox}
\caption{Performance up to the SFT stage}
\label{tab:results_sft}
\end{subtable}

% --- 子表 (b) --- 将 \textwidth 改为 0.9\textwidth
\begin{subtable}{0.9\textwidth}
\centering
% 将 width 同样改为 0.9\textwidth
\begin{adjustbox}{width=0.9\textwidth,center}
\begin{tabular}{@{} l ccc cccc c @{}}
\toprule
\multirow{2}{*}{\textbf{Method}} & \multicolumn{3}{c}{\textbf{Single-Hop QA}} & \multicolumn{4}{c}{\textbf{Multi-Hop QA}} & \multirow{2}{*}{\textbf{Avg.}} \\
\cmidrule(lr){2-4} \cmidrule(lr){5-8}
& \textbf{NQ} & \textbf{TriviaQA} & \textbf{PopQA} & \textbf{HotpotQA} & \textbf{2Wiki} & \textbf{MuSiQue} & \textbf{Bamboogle} & \\
\midrule
\multicolumn{9}{c}{\textit{Youtu-LLM-2B-Base}} \\
\midrule
SFT \& RL* & 48.0 & 58.2 & 35.2 & 53.8 & 45.0 & 30.2 & 52.0 & 46.1 \\
FE-SFT \& RL* & 49.4 & 60.0 & 38.2 & 55.4 & 42.0 & 29.8 & \underline{54.4} & 47.0 \\
State-SFT \& RL$\dagger$ & 48.2 & 58.8 & 40.8 & 55.2 & 40.8 & 31.4 & \underline{54.4} & 47.1 \\
IWM \& RL & 50.0 & 62.2 & 38.8 & 55.0 & \underline{46.0} & \underline{33.2} & 53.6 & 48.4 \\
\midrule
\multicolumn{9}{c}{\textit{Youtu-LLM-2B-Base (WM-AMT, ours)}} \\
\midrule
SFT \& RL & 50.2 & 59.8 & 35.6 & 53.6 & \underline{46.0} & 25.4 & 45.6 & 45.2 \\
FE-SFT \& RL & \underline{52.6} & \underline{63.2} & 40.8 & \underline{57.0} & 44.4 & 32.6 & 52.8 & \underline{49.1} \\
State-SFT \& RL$\ddagger$ & 52.8 & \textbf{64.4} & \textbf{42.8} & 54.6 & 42.2 & 30.4 & 53.6 & 48.7 \\
\midrule
FE-SFT \& FC-RL & \textbf{53.4} & 62.6 & \underline{42.6} & \textbf{59.0} & \textbf{46.2} & \textbf{34.2} & \textbf{56.0} & \textbf{50.6} \\
\bottomrule
\end{tabular}
\end{adjustbox}
\caption{Performance with the RL stage (the final performance)}
\label{tab:results_rl}
\end{subtable}
\caption{Main results on search: performance on 7 QA datasets. \textit{SFT} uses only \textit{think} and \textit{action} segments, excluding world model block injections. \textit{RL} denotes the standard GRPO algorithm. \textit{State-SFT} corresponds to (2) and (3) in the Baselines; these models predict future state transitions but omit verbalized Q-values. These denotations remain the same as in Table \ref{aime_results}. The \textbf{bold text} represents the best, while \underline{underlined} represents the second best.}
\label{search_results}
\end{table*}

\subsection{Main results}
\label{subsec: main_results}
\paragraph{Necessity of our three phases.} \textbf{(1) WM-AMT}: At the SFT stage, replacing standard agentic mid-training with our WM-AMT improves the average score under the same SFT method (e.g., in Table \ref{search_results}, 38.2 $\rightarrow$ 39.9 for SFT and 37.4 $\rightarrow$ 41.8 for FE-SFT). We provide a case study in Appendix \ref{subsec: gap_example}: under the same FE-SFT format, our WM-AMT model generates more coherent, causally accurate world model blocks, while the base model fails. These results indicate that FE-SFT is substantially effective when the model has internalized future-dynamics priors via WM-AMT. \textbf{(2) FE-SFT}: Under our WM-AMT model, our FE-SFT outperforms standard SFT (e.g. 39.9 $\rightarrow$ 41.8 in Table \ref{search_results}), showing that explicitly eliciting structured foresight is beneficial after the world-model capability is established. \textbf{(3) FC-RL}: In RL-stage comparisons, our FE-SFT \& FC-RL achieves the best overall average (50.6 in Table \ref{search_results}, mean@30: 29.5 and pass@30: 60.0 in Table \ref{aime_results}), outperforming all the other baselines. 

\paragraph{Stage dependency.}
The overall results exhibit a progressive dependency: WM-AMT provides a capability foundation, FE-SFT activates this capability in a structured format, and FC-RL optimizes end-task behavior with grounded foresight. Removing or replacing any stage weakens the final performance, confirming the need for the full pipeline. 

% We compare our method with State-Only Prediction baselines ($\dagger$ and $\ddagger$).
\paragraph{Comparison with other world modeling methods.}  In Table \ref{search_results}, the highest average score of state-only baselines is 48.7; while in Table \ref{aime_results}, it achieves a pass@30 of 56.7 and a mean@30 of 27.7. By adding the verbalized $Q$-value and applying our FC-RL to jointly optimize the policy and its internalized world model, our final model (WM-AMT \& FE-SFT \& FC-RL) increases these scores to 50.6 in Table \ref{search_results}, 60.0 (pass@30) and 29.5 (mean@30) in Table \ref{aime_results}, respectively. It proves that simply predicting future states is not enough for world modeling. Instead, explicitly adding the confidence estimation and utilizing the calibrated confidence to guide the policy are crucial for making better decisions. Notably, IWM (+RL) achieves strong performance on Youtu-LLM-2B-Base, but our final model still shows a significant and consistent improvement.

\paragraph{Efficacy on complex multi-hop reasoning.} Another notable fact is that our proposed paradigm yields particularly pronounced improvements on complex, multi-step reasoning tasks in Table \ref{search_results}. Up to the SFT stage, the transition from \textit{SFT} to \textit{FE-SFT} under our WM-AMT model brings a significant improvement (MuSiQue: 20.6 $\rightarrow$ 25.6, 2WikiMultiHopQA: 31.2 $\rightarrow$ 35.8). Our final model also demonstrates substantial superiority on multi-hop datasets compared to the other baselines, e.g., HotpotQA (59.0), MuSiQue (34.2) and Bamboogle (56.0). It indicates that the internalized world model and foresight-conditioned planning are especially critical for tasks requiring long-horizon reasoning, whereas simple factual retrieval (single-hop) relies less heavily on future simulation.

\begin{table*}[t]
\centering
% \scriptsize  % 如果缩放后觉得字稍微有点大，可以取消这一行的注释

% ==================== 子表 (a) SFT Stage ====================
\begin{subtable}{0.85\textwidth}  % 子标题限制在 0.8 宽度
\centering
\begin{adjustbox}{width=0.85\textwidth,center} % 强制表格内容整体缩放到 0.8 宽度且居中
\begin{tabular}{@{} l cccc | cccc @{}} % 改为普通的 tabular，c 保证数据居中且绝对不换行
\toprule
\multirow{2}{*}{\textbf{Method}} & \multicolumn{4}{c}{\textbf{mean@30}} & \multicolumn{4}{c}{\textbf{pass@30}} \\
\cmidrule(lr){2-5} \cmidrule(lr){6-9}
& \textbf{AIME 24} & \textbf{AIME 25} & \textbf{AIME 26} & \textbf{Avg.} & \textbf{AIME 24} & \textbf{AIME 25} & \textbf{AIME 26} & \textbf{Avg.} \\
\midrule
\multicolumn{9}{c}{\textit{Youtu-LLM-2B-Base}} \\
\midrule
SFT* & 9.1 & 9.9 & 8.9 & 9.3 & 46.7 & \underline{33.3} & \underline{40.0} & 40.0 \\
FE-SFT* & 12.0 & 9.9 & 8.0 & 10.0 & \underline{53.3} & \textbf{36.7} & \underline{40.0} & \underline{43.3} \\
IWM & \textbf{13.0} & \underline{10.6} & \textbf{10.2} & \underline{11.3} & \textbf{56.7} & \underline{33.3} & \textbf{43.3} & \textbf{44.4} \\
\midrule
\multicolumn{9}{c}{\textit{Youtu-LLM-2B-Base (WM-AMT, ours)}} \\
\midrule
SFT & 10.9 & \underline{10.6} & 8.8 & 10.1 & \textbf{56.7} & \textbf{36.7} & 33.3 & 42.2 \\
FE-SFT & \underline{12.2} & \textbf{12.2} & \underline{9.7} & \textbf{11.4} & \underline{53.3} & \textbf{36.7} & \underline{40.0} & \underline{43.3} \\
\bottomrule
\end{tabular}
\end{adjustbox}
\caption{Performance up to the SFT stage}
\end{subtable}

% ==================== 子表 (b) RL Stage ====================
\begin{subtable}{0.9\textwidth}  % 子标题限制在 0.8 宽度
\centering
\begin{adjustbox}{width=0.9\textwidth,center} % 强制表格内容整体缩放到 0.8 宽度且居中
\begin{tabular}{@{} l cccc | cccc @{}} % 改为普通的 tabular
\toprule
\multirow{2}{*}{\textbf{Method}} & \multicolumn{4}{c}{\textbf{mean@30}} & \multicolumn{4}{c}{\textbf{pass@30}} \\
\cmidrule(lr){2-5} \cmidrule(lr){6-9}
& \textbf{AIME 24} & \textbf{AIME 25} & \textbf{AIME 26} & \textbf{Avg.} & \textbf{AIME 24} & \textbf{AIME 25} & \textbf{AIME 26} & \textbf{Avg.} \\
\midrule
\multicolumn{9}{c}{\textit{Youtu-LLM-2B-Base}} \\
\midrule
SFT \& RL* & \textbf{35.1} & 24.3 & 24.6 & 28.0 & 56.7 & 46.7 & 53.3 & 52.2 \\
FE-SFT \& RL* & 31.9 & \underline{26.4} & 25.7 & 28.0 & 60.0 & 46.7 & 46.7 & 51.1 \\
State-SFT \& RL$\dagger$ & 32.2 & 22.1 & 27.0 & 27.1 & 60.0 & 50.0 & \textbf{60.0} & 56.7 \\
IWM \& RL & 33.2 & 24.9 & \textbf{27.6} & \underline{28.6}  & \textbf{66.7} & 50.0 & \underline{56.7}  & \underline{57.8} \\
\midrule
\multicolumn{9}{c}{\textit{Youtu-LLM-2B-Base (WM-AMT, ours)}} \\
\midrule
SFT \& RL & 34.2 & 23.4 & 25.9 & 27.8 & \underline{63.3} & \underline{53.3} & 53.3 & 56.6 \\
FE-SFT \& RL & 33.6 & 25.6 & 26.0 & 28.4 & 60.0 & \underline{53.3} & 50.0 & 55.5 \\
State-SFT \& RL$\ddagger$ & 33.2 & 24.8 & 25.0 & 27.7 & \underline{63.3} & \textbf{56.7} & 50.0 & 56.7 \\
\midrule
FE-SFT \& FC-RL & \underline{34.6} & \textbf{26.8} & \underline{27.1} & \textbf{29.5} & \textbf{66.7} & \underline{53.3} & \textbf{60.0} & \textbf{60.0} \\
\bottomrule
\end{tabular}
\end{adjustbox}
\caption{Performance with the RL stage (the final performance)}
\end{subtable}

\caption{Main results on mathematical reasoning. The \textbf{bold text} and \underline{underlined} represent the same as in Table \ref{search_results}.}
\label{aime_results}
\end{table*}

\paragraph{Qualitative analysis of world model blocks. } We provide examples in Appendix \ref{web_example} and \ref{math_example} to analyze the world model blocks. Two key capabilities emerge: 
(1) \textbf{Long-horizon planning coherence}: these blocks establish 
reasoning frameworks that persist across multiple turns, even when intermediate errors require correction; and (2) \textbf{Calibrated confidence quantification}: confidence estimates correlate meaningfully with 
the actual task success, decreasing upon error detection and remaining low for 
inherently challenging problems.

Additional ablation studies, sensitivity analyses on $\omega$ in FC-RL, and analyses on the response length are provided in Appendix \ref{subsec:sensitivity}, \ref{subsec:ablation} and \ref{subsec:res_length}.

\section{Conclusion}
We address the limitations of reactive LLM agents by internalizing future-aware planning into the policy. We identify the \textit{format-capability gap}, wherein post-training on look-ahead formats merely teaches structural mimicry instead of grounded prediction. To bridge this gap, we propose a unified three-stage training paradigm consisting of WM-AMT, FE-SFT and FC-RL. Evaluations across search and mathematical reasoning show the superiority of our approach over other training baselines. Our results underscore that effective internal world modeling in LLM agents demands a capability-first paradigm to achieve truly grounded and calibrated foresight.

\section*{Contributions}

\paragraph{Authors}
Xuan Zhang\textsuperscript{\rm 1,2,3}\quad Zhijian Zhou\textsuperscript{\rm 1,2,3}\quad LingFeng Qiao \textsuperscript{\rm 3} \quad Yulei Qin\textsuperscript{\rm 3}
Ke Li\textsuperscript{\rm 3}\quad Xing Sun\textsuperscript{\rm 3} \\
Xiaoyu Tan\textsuperscript{\rm 3 $\dagger$} \quad Chao Qu\textsuperscript{\rm 1 $\dagger$} \quad Yuan Qi\textsuperscript{\rm 1,2 $\dagger$} 

\paragraph{Affiliations}
\textsuperscript{\rm 1}Fudan University\quad \textsuperscript{\rm 2}Shanghai Innovation Institute\quad \textsuperscript{\rm 3}Tencent Youtu Lab

\paragraph{$^\dagger$Corresponding Authors}
Xiaoyu Tan \quad Chao Qu \quad Yuan Qi

\paragraph{Acknowledgments}
We greatly thank the VeRL \citep{Sheng_2025} and the VeRL-agent \citep{feng2025groupingroup} communities for their implementation of various RL training and inference frameworks.
% Acknowledgments should be placed at the end of the paper, before the bibliography. This section can be used to thank individuals, organizations, or funding sources that contributed to the research.

\setcitestyle{numbers,square}
%% \setcitestyle{square,numbers,comma}
%% \bibliographystyle{unsrt}
%% \bibliographystyle{plainnat}

\bibliography{youtu_bib}

\appendix
\section{Addition descriptions for our method}
\label{add_alg}
\subsection{World model agentic mid-training}
Our algorithmic description for WM-AMT is shown in Algorithm \ref{alg:wm_amt}.
\begin{algorithm*} 
    \caption{World Model Agentic Mid-Training (WM-AMT)} 
    \label{alg:wm_amt} 
    \begin{algorithmic}
        \REQUIRE Raw trajectory dataset $\mathcal{D}_{\text{raw}} = \{ \tau_i \}_{i=1}^N$, policy $\theta$, summarization function $\Phi$.
        \STATE Initialize the augmented dataset $\mathcal{D}_{\text{WM-AMT}} \leftarrow \emptyset$.
        
        \FOR{each multi-turn interaction trajectory $\tau = (s_1, a_1, \dots, s_T, a_T)$ in $\mathcal{D}_{\text{raw}}$}
            \STATE Sample a set of indices $\mathcal{I} \sim \mathcal{U}(1, T-1)$.
            \FOR{each step $t \in \mathcal{I}$}
                \STATE Generate a structured world model block $z_t = \Phi(s_t, \tau_{>t})$, composed of $z_t = [z_t^{(\text{traj})} \oplus z_t^{(\text{eval})}]$.
                \STATE Estimate a Q-value confidence score $q \in [0, 100]\%$ via an LLM to represent task success probability, and incorporate it into $z_t^{(\text{eval})}$.
            \ENDFOR
            \STATE Inject generated $\{z_t\}_{t \in \mathcal{I}}$ into $\tau$, yielding an augmented trajectory $\tilde{\tau} = (\dots, s_t, z_t, a_t, s_{t+1}, \dots)$.
            \STATE Add $\tilde{\tau}$ to the augmented dataset $\mathcal{D}_{\text{WM-AMT}}$.
        \ENDFOR
        
        \STATE \textbf{Optimization Phase:}
        \WHILE{not converged}
            \STATE Sample a batch of trajectories $\tilde{\tau} \sim \mathcal{D}_{\text{WM-AMT}} \cup \{ \tau | \tau \in \mathcal{D}_{\text{raw}} \ \wedge \tau  \notin \mathcal{D}_{\text{WM-AMT}}\}$.
            \STATE Compute the autoregressive negative log-likelihood loss: 
            $\mathcal{L}_{\text{WM-AMT}}(\theta) = - \mathbb{E}_{\tilde{\tau}} \left[ \sum_{i=1}^{|\tilde{\tau}|} \log P_\theta(u_i | u_{<i}) \right]$.
            \STATE Update the policy $\theta$ to minimize $\mathcal{L}_{\text{WM-AMT}}(\theta)$.
        \ENDWHILE
        
        \ENSURE Mid-trained policy model $\theta$.
    \end{algorithmic} 
\end{algorithm*}

\subsection{Foresight-conditioned reinforcement learning}
Our algorithmic description for FC-RL is shown in Algorithm \ref{alg:fg_rl}.

\begin{algorithm*} 
    \caption{Foresight-Conditioned Reinforcement Learning (FC-RL)} 
    \label{alg:fg_rl} 
    \begin{algorithmic}
        \REQUIRE Policy model $\pi_\phi$, prompt dataset $\mathcal{D}$, group size $N$, balance weight $\omega$.
        
        \FOR{each prompt $x$ in $\mathcal{D}$}
            \STATE \textbf{\% Phase 1: Group Trajectory Generation}
            \FOR{$i=1$ to $N$}
                \STATE Generate trajectory $\tau_i$ autoregressively via an interaction loop:
                \WHILE{task not terminated at step $t$}
                    \STATE Generate World Model Span $\mathcal{T}_{\text{WM}}$: imagine and predict keywords $K_t^{(i)}$ and confidence $q_t^{(i)}$.
                    \STATE Generate Policy Span $\mathcal{T}_{\text{Policy}}$: think and execute action $a_t^{(i)}$.
                \ENDWHILE
                \STATE Obtain terminal correctness indicator $\mathbf{1}_{y_i \equiv y_i^*}$ and episode reward $R^E_i$.
            \ENDFOR
            
            \STATE \textbf{\% Phase 2: Grounding \& Calibration (Group-wise)}
            \STATE Compute global advantage $A^E(\tau_i)$ by normalizing $\{R^E_i\}_{i=1}^N$ across the group.
            \FOR{$i=1$ to $N$}
                \FOR{each step $t$ in $\tau_i$}
                    \STATE \textit{\# 2.1 World Model Optimization (Foresight grounding and calibration)}
                    \STATE Verification of grounding: $R_{\text{ground}}^{i,t} = R_{\text{match}}(K_t^{(i)}, \tau_{>t}^{(i)})$.
                    \STATE Brier score penalty: $R_{\text{calib}}^{i, t} = \mathbf{1}_{y_i \equiv y_i^*} - (q_t^{(i)} - \mathbf{1}_{y_i \equiv y_i^*})^2$.
                    \STATE Combined WM reward: $R^{\text{WM}}(\tau_i^t) = R_{\text{calib}}^{i, t} + R_{\text{ground}}^{i,t}$.
                    \STATE Compute WM advantage $A(\tau[\text{world\_model}]_i^t)$ by normalizing $\{R^{\text{WM}}(\tau_j^t)\}_{j=1}^N$.
                    
                    \STATE \textit{\# 2.2 Policy Optimization (Execution success)}
                    \STATE Compute step advantage $A^S(\tau_i^t)$ by normalizing confidence scores $\{q_t^{(j)}\}_{j=1}^N$.
                    \STATE Combined policy advantage: $A(\tau[\text{policy}]_i^t) = A^E(\tau_i) + \omega \cdot A^S(\tau_i^t)$.
                \ENDFOR
            \ENDFOR
            
            \STATE \textbf{\% Phase 3: Policy Update}
            \STATE Apply $A(\tau[\text{world\_model}]_i^t)$ to optimize the $\mathcal{T}_{\text{WM}}$ spans.
            \STATE Apply $A(\tau[\text{policy}]_i^t)$ to optimize the $\mathcal{T}_{\text{Policy}}$ spans.
        \ENDFOR
        
        \ENSURE Optimized policy model $\pi_\phi$.
    \end{algorithmic} 
\end{algorithm*} 

\section{Data construction}
\label{app:prompts}
\subsection{World model agentic mid-training}

To construct the structured world model Block $z_t$ during the World Model Agentic Mid-Training (WM-AMT) phase, we employ an LLM acting as the summarization function $\Phi(\cdot)$. The prompt enforces abstract planning to prevent future-state data leakage and explicitly requires confidence estimation for $Q$-value modeling. The exact prompt template is provided below:

\begin{tcolorbox}[
    colback=SeaGreen!10!CornflowerBlue!10,
    colframe=RoyalPurple!55!Aquamarine!100!,
    title=Prompt Template for Summarization ($\Phi$), % 标题
    fonttitle=\bfseries,        % 标题字体加粗
    boxrule=0.8pt,              % 边框粗细
    arc=3pt                     % 圆角弧度
]
\begin{verbatim}
Act as an agent's "World Model". Based on the **FUTURE GROUND TRUTH** (the 
verified success path), generate the internal reasoning.

**CRITICAL RULES:**
1. **NO SPOILERS:** Plan the future steps using *abstract placeholders* (e.g., 
   "expect to find a candidate name"), NOT specific entities found in the future.
2. **BE SPECIFIC:** In the Action Plan, specify exact keywords/intents, 
   avoiding generic phrases like "Proceed".

**OUTPUT FORMAT (Markdown Only):**

## World Model Simulation
- **Trajectory:** Step-by-step roadmap to the answer based on the Future Truth.
- **Gap:** Why the current state is insufficient.
- **Confidence:** [0-100]%. (High=Direct path, Low=Complex/Trial-error). 
  Brief reason.

## Immediate Action Plan
- The precise intent for the *immediate next action* in the 
  Ground Truth.

**DATA:**
[Current Obs]:
{current_obs}
[FUTURE TRUTH]:
{future_traj}
\end{verbatim}
\end{tcolorbox}

We leverage DeepSeek V3.1 \citep{deepseekai2025deepseekv3technicalreport} as the foundational data generator for World Model Agentic Mid-Training. The model is queried using the above prompt, with generation parameters set to a temperature of 0.6, a \textit{top-p} of 0.95, and a maximum response length of 32768 tokens.

The original agentic trajectory data consists of 200B tokens of high-quality trajectory data, including 25B Agentic-CoT trajectories, 20B mathematical trajectories, 70B code execution trajectories, 60B Deep Research trajectories, and 25B other trajectories (e.g., tool using, function calling, and planning).

In WM-AMT, we augment the original agentic trajectory data by inserting an explicit \textit{world model block} into a random split of multi-turn interaction trajectories, which are primarily concentrated in Deep Research trajectories and mathematical trajectories. After our data augmentation, we obtained roughly $\sim 62$B tokens for Deep Research trajectories and $\sim21$B for mathematics, whereas the remaining trajectory categories remained unaltered.

Some training samples constructed by this way are shown as follows:
% \clearpage
% \onecolumn
\begin{databox}{(A) A \textit{Deepresearch} Trajectory Sample (Trajectory + World Model Block)}
\begin{fieldbox}{User Question}
Under which record label was the 17th studio album of the Ivory Coast reggae musician known as the 'Bob Marley of Africa' released in 2015, who also performed at the Street Mode Festival in Greece and collaborated with Harry Roselmack on a tribute song in 2014?

You start with one or more cycles of (thinking about which tool to use $\rightarrow$ performing tool call $\rightarrow$ waiting for tool response), and end with (thinking about the answer $\rightarrow$ answer of the question).

The thinking processes, tool calls, tool responses, and answer are enclosed within their tags.

There could be multiple thinking processes, tool calls, tool call parameters and tool response parameters.

When you finally reach the conclusion, put your final answer enclosed with $<$answer$>$ your final answer here $<$/answer$>$.
\end{fieldbox}

\begin{fieldbox}{Turn 1: World Model Block}
**Trajectory:** \par  1. Identify the musician via search using the descriptive title "Ivory Coast reggae musician known as the 'Bob Marley of Africa'".\par  2. Confirm the artist's name (Alpha Blondy) from the initial search results.\par  3. Execute a refined search for the specific details: the 17th studio album (2015 release), the festival performance, and the collaboration.\par  4. Extract the album title and record label from the detailed results about the 2015 release.\par  5. Synthesize the information to provide the final answer.\par **Gap:** The current observation only contains the user's question. No information about the artist's identity, discography, or specific events is available to answer the query directly.\par **Confidence:** 85\%. The question provides multiple unique, verifiable identifiers (specific album number, year, festival, collaboration) that create a direct path to the answer through targeted searches. The primary complexity is confirming the artist's name first.\par\par **Intent:** Identify the correct musician by searching for the unique descriptive title."
\end{fieldbox}

\begin{fieldbox}{Turn 1: Action}
search({"query": ["Ivory Coast reggae musician known as the 'Bob Marley of Africa'", "17th studio album of the 'Bob Marley of Africa' released in 2015", "Street Mode Festival Greece reggae musician Ivory Coast", "Ivory Coast reggae musician collaborated with Harry Roselmack 2014"]})
\end{fieldbox}

\begin{fieldbox}{Turn 1: Feedback}
Results for query "Ivory Coast reggae musician known as the 'Bob Marley of Africa'": ......

Results for query "17th studio album of the 'Bob Marley of Africa' released in 2015":  ......

Results for query "Street Mode Festival Greece reggae musician Ivory Coast": ......

Results for query "Ivory Coast reggae musician collaborated with Harry Roselmack 2014":  ......
\end{fieldbox}

\begin{fieldbox}{Turn 2: World Model Block}
    **Trajectory:** \par   1. Confirm the candidate's identity as the verified "Bob Marley of Africa" from the initial search results.\par   2. Search specifically for the candidate's 17th studio album released in 2015 using their confirmed name.\par   3. Search specifically for the candidate's performance at the Street Mode Festival using their confirmed name.\par   4. Search specifically for the candidate's collaboration with Harry Roselmack in 2014 using their confirmed name.\par   5. Extract the album title, festival performance year, and collaboration details from the targeted results.\par \par **Gap:** The current queries using descriptive placeholders ("Bob Marley of Africa") instead of the confirmed name are returning irrelevant results about Bob Marley himself and failing to find festival/collaboration specifics due to low specificity.\par \par **Confidence:** 85\%. The path is direct once the candidate's identity is confirmed, but requires precise query formulation to bypass noise about the original Bob Marley's discography and vague festival/collaboration references.\par \par **Immediate Action Plan** \par Execute a search with the exact query: "Alpha Blondy 17th studio album 2015" to identify the album title and label.\par Execute a search with the exact query: "Alpha Blondy Street Mode Festival" to confirm the performance year.\par Execute a search with the exact query: "Alpha Blondy Harry Roselmack 2014" to identify the collaboration project.
\end{fieldbox}

\begin{fieldbox}{Turn 2: Think}
The search results indicate that the musician referred to as the "Bob Marley of Africa" is Alpha Blondy from Ivory Coast. However, the results do not provide specific details about his 17th studio album released in 2015, his performance at the Street Mode Festival in Greece, or his collaboration with Harry Roselmack in 2014. \par To gather more precise information, I will refine the search queries to focus on Alpha Blondy's 17th studio album and his collaborations.
\end{fieldbox}

\begin{fieldbox}{Turn 2: Action}
search({"query": ["Alpha Blondy 17th studio album 2015", "Alpha Blondy Street Mode Festival Greece", "Alpha Blondy collaboration with Harry Roselmack 2014"]})
\end{fieldbox}

\begin{fieldbox}{Turn 2: Feedback}
Results for query "Alpha Blondy 17th studio album 2015" ......

Results for query "Alpha Blondy Street Mode Festival Greece" ......

Results for query "Alpha Blondy collaboration with Harry Roselmack 2014" ......
\end{fieldbox}

\begin{fieldbox}{Turn 3: Think}
The search results provide the following information:\par1. **Alpha Blondy's 17th Studio Album (2015)**: The album is titled *Positive Energy*, released on May 18, 2015, under the label **Wagram Music** (as mentioned in the [World A Reggae article](https:....../) and [Discogs entry](https:......)).\par\par
2. **Performance at Street Mode Festival (Greece)**: Alpha Blondy performed at the Street Mode Festival in Thessaloniki, Greece, in 2016 (as per [Setlist.fm](https:......) and the festival's [history page](https:....../)).\par\par
3. **Collaboration with Harry Roselmack (2014)**: Alpha Blondy collaborated with Harry Roselmack and other artists on the song *On n'oublie pas* in 2014, a tribute song (as mentioned in [Tanya St Val's profile](https:......) and [TikTok references](https:......)).\par\par
\#\#\# Final Answer:\par$<$answer$>$ \par The 17th studio album of Alpha Blondy, the "Bob Marley of Africa," released in 2015 was *Positive Energy*, under the record label **Wagram Music**. He performed at the Street Mode Festival in Greece in 2016 and collaborated with Harry Roselmack on the tribute song *On n'oublie pas* in 2014.\par $<$/answer$>$
\end{fieldbox}

\end{databox}

\begin{databox}{(B) A \textit{Mathematical} Trajectory Sample (Trajectory + World Model Block)}

\begin{fieldbox}{User Question}
A bear eats up to 90 pounds of food per day to prepare for hibernation. If Victor weighs 126 pounds, how many "Victors" worth of food would a bear eat in 3 weeks?
\end{fieldbox}

\begin{fieldbox}{Turn 1: World Model Block}
**Reasoning Flow:** \par   1. Calculate total days in 3 weeks (3 × 7 = 21 days)\par   2. Compute total food consumed over 21 days (daily consumption × 21)\par   3. Determine the ratio of total food to Victor's weight (total\_food / victor\_weight)\par   4. Output the final unitless count representing "Victors" worth\par \par  **Gap:** The current state only provides the problem statement without numerical computation. The solution requires executing the mathematical operations to transform inputs (90 lbs/day, 126 lbs, 3 weeks) into the requested ratio.\par \par  **Confidence:** 100\%. The problem is straightforward with a direct mathematical path: multiplication followed by division. No branching logic or approximations are needed.\par \par **Action to Take** \par  Calculate total days in 3 weeks: Multiply 3 by 7\par  Compute total food consumption: Multiply daily food (90) by total days\par  Derive the Victor ratio: Divide total food by Victor's weight (126)\par  Output the final value as the answer

\end{fieldbox}

\begin{fieldbox}{Turn 1: Think}
Alright, let's break this problem down step by step.\par \par **Step 1:** Calculate how much food a bear eats in one day.\par This is given as 90 pounds.\par \par **Step 2:** Calculate how much food a bear eats in 3 weeks.\par There are 7 days in a week, so in 3 weeks, there are \(3 \times 7 = 21\) days.\par So, the total food a bear eats in 3 weeks is \(21 \times \text{(food eaten in one day)}\).\par \par **Step 3:** Calculate how many "Victors" worth of food a bear would eat in 3 weeks.\par This is calculated by dividing the total food a bear eats in 3 weeks (from Step 2) by Victor's weight.\par \par Let's now calculate these using the code interpreter.
\end{fieldbox}

\begin{fieldbox}{Turn 1: Action}
\# Given data\par food\_per\_day = 90  \# pounds\par victor\_weight = 126  \# pounds\par \# Step 2: Calculate total food in 3 weeks\par total\_food\_3\_weeks = 21 * food\_per\_day\par \# Step 3: Calculate how many "Victors" worth of food in 3 weeks\par victors\_worth = total\_food\_3\_weeks / victor\_weight\par victors\_worth
\end{fieldbox}

\begin{fieldbox}{Turn 1: Feedback}
15.0
\end{fieldbox}

\begin{fieldbox}{Turn 2: Think}
The bear would eat the equivalent of ( \boxed{15} ) "Victors" worth of food in 3 weeks.
\end{fieldbox}

\end{databox}
% \clearpage
% \twocolumn

\subsection{Format-eliciting supervised fine-tuning}
The prompt and generation configurations for foresight block construction in FE-SFT is as the same as the prompt in WM-AMT, with only additional formatting applied. Particularly, the seperate prompt for extracting keywords from the foresight block is as follows:

% \begin{table*}
\begin{tcolorbox}[
    colback=SeaGreen!10!CornflowerBlue!10,
    colframe=RoyalPurple!55!Aquamarine!100!,
    title=Prompt Template for Keywords Extraction, % 标题
    fonttitle=\bfseries,        % 标题字体加粗
    breakable,                  % 允许跨页断行
    boxrule=0.8pt,              % 边框粗细
    arc=3pt                     % 圆角弧度
]
\begin{verbatim}
Please identify the top 5 keywords 
from the text below. Provide only the
keywords, separated by commas.
\end{verbatim}
\end{tcolorbox}
% \end{table*}

We also leverage DeepSeek V3.1 for extracting keywords, with generation parameters set to a temperature of 0.6, a \textit{top-p} of 0.95, and a maximum response length of 4096 tokens.

We constructed $\sim 4k$ trajectories for the search task, reformulated into $\sim 13k$ single-turn samples to align with the verl-agent framework for SFT. For the mathematical reasoning task, we curated $\sim 2k$ trajectories specifically for SFT.

Some training samples constructed by this way are shown as follows:

\begin{databox}{(A) An example of \textit{Search} task}
\begin{fieldbox}{User Question}
Answer the given question step by step. Start by drawing a plan inside $<$imaginary$>$Your imagination, which predicts the subsequent steps and plans you will take$<$/imaginary$>$. Then, you should summarize your plans with words/phrases inside $<$keyword$>$ and $<$/keyword$>$, and evaluate your plans inside $<$analysis$>$Your analysis on your plans$<$/analysis$>$ and $<$confidence$>$Your confidence$<$/confidence$>$. Following the above imagination and plan, you can reason inside $<$think$>$Your thought$<$/think$>$. Evert time you get new information, you should plan and reason again. If you lack knowledge, search using $<$search$>$Your query$<$/search$>$ and it will return the top searched results in $<$information$>$Your information$<$/information$>$. You can search multiple times as needed and each search should only search for one information about only one entity. For questions concerning more than one entities, break them down into steps and search for each piece of information. Finally, if you find no further external knowledge needed, you can directly provide the answer in $<$answer$>$Your answer$<$/answer$>$ without detailed explanations. For example, $<$answer$>$Beijing$<$/answer$>$. Let's search step by step.\par
Question: \textcolor{blue}{In what year was the clown who, together with Vladimir Olshansky, founded the clown theatre comany Litsedei born?} \par
Below is the interaction history:
\end{fieldbox}

\begin{fieldbox}{Turn 1: Action}
$<$imaginary$>$The path involves two sequential information retrieval steps. First, identify the name of the co-founder of the Litsedei clown theatre company. Second, locate the specific birth year for that identified individual. The process is a straightforward lookup once the correct entity is found. The current state lacks the identity of the co-founder and, consequently, the data point (birth year) associated with that individual.$<$/imaginary$>$\par
$<$keyword$>$information, retrieval, co-founder, Litsedei, theatre$<$/keyword$>$\par
$<$analysis$>$The task is a direct factual query with a high probability of finding the answer in standard biographical sources. The only minor uncertainty is in the initial search correctly identifying the co-founder's name.$<$/analysis$>$\par
$<$confidence$>$95\%$<$/confidence$>$\par
$<$think$>$I need to find information about the clown who co-founded the Litsedei clown theatre company with Vladimir Olshansky, and specifically determine their birth year. Let me start by searching for information about Litsedei theatre and its founders.$<$/think$>$\par
$<$search$>$Litsedei clown theatre company founders Vladimir Olshansky$<$/search$>$
\end{fieldbox}

\begin{fieldbox}{Turn 1: Observation}
$<$information$>$\par
Doc 1: "Vladimir Olshansky" \par
Vladimir Olshansky Vladimir Olshansky (9 August 1947 in St. Petersburg, Russia) is a performing artist, director, composer, sculptor, graduate of the Moscow State Circus School, together with clown Slava Polunin, he is the founder of the clown theatre company Litsedei. He creates numerous one man shows and performs at the Leningrad's State Variety Theatre, In parallel he studies directing at the Leningrad's State Theatre and Cinema Institute. At the end of 1980s Vladimir moves to New York, where he becomes a supervisor hospital clown, working with ...\par
 Doc 2: "Vladimir Olshansky"\par
 2017 "Manuale di clownterapia", Rome, Italy Edizione Dino Audino Ann-Kathrin Soder "Arena Spots" Erlangen, Germany 2/06/07 Vladimir Olshansky Vladimir Olshansky (9 August 1947 in St. Petersburg, Russia) is a performing artist, director, composer, sculptor, graduate of the Moscow State Circus School , together with clown Slava Polunin, he is the founder of the clown theatre company Litsedei. He creates numerous one man shows and performs at the Leningrad's State Variety Theatre , In parallel he studies directing at the Leningrad's State Theatre and Cinema Institute  At the end ... \par
 Doc 3: "Vladimir Olshansky"\par
 Engibarov who makes a great impression on him inspiring his dream to create, one day, his own clown-theatre. Engibarov was the first clown to combine the arts of both the circus and the theatre. During the period, from the early 1970s to the 1980s, Olshansky collaborates with the Russian directors: Lev Stukalov (now artistic director of St. Petersburg's "Our Theatre" , Vlad Druzinin and Yuri Gerzman, Victor Charitonov, the artist and papet maker Natalia Lazareva, set designers: Simon Pastuch, Emil Kapelush, caricaturist Victor Bogorad, and creates his sketch "The Swan" becoming a winner at ... \par
$<$/information$>$
\end{fieldbox}

\begin{fieldbox}{Turn 2: Action}
    $<$imaginary$>$The trajectory is complete. The agent has successfully identified the target individual (the co-founder) and now needs to retrieve a specific biographical fact about that individual. The next step is to execute a targeted search for that fact. The information retrieved will be scanned to locate the specific data point, which will conclude the task.\par The current state is insufficient because the agent possesses the identity of the co-founder but lacks the specific biographical data (birth year) required to answer the final question.$<$/imaginary$>$\par $<$keyword$>$agent, target, biographical, fact, search$<$/keyword$>$\par $<$analysis$>$The path is direct and low-risk. The target entity is clearly identified, and the required data point is a standard, well-documented fact that is highly likely to be present in the top results of a precise search.$<$/analysis$>$\par $<$confidence$>$95\%$<$/confidence$>$\par $<$think$>$Great! I found that the co-founder of Litsedei clown theatre company with Vladimir Olshansky is clown Slava Polunin. Now I need to find Slava Polunin's birth year.$<$/think$>$\par $<$search$>$Slava Polunin birth year$<$/search$>$
\end{fieldbox}

\begin{fieldbox}{Turn 2: Observation}
$<$information$>$\par
Doc 1: "Slava Polunin"\par Slava Polunin Vyacheslav Ivanovich Slava Polunin PAR (born 12 June 1950) is a Russian performance artist and clown. He is the creator of the stage spectacles "Asisyai-revue", "Slava's Snowshow" and "Diabolo". Polunin was born in the town of Novosil, Oryol Oblast, Russia, into the family of a shop assistant. He was successful in his school theatre, imitating Charlie Chaplin, but was refused entry to the Leningrad Theater Institute due to poor pronunciation. After a few years' study at an engineering school, he graduated from the Leningrad Institute for Soviet Culture, where ... \par
Doc 2: "Slava Polunin"\par many prestigious awards including Gold Angle (Scotland), Gold Nose (Spain), Triumph (Russia) and the Laurence Olivier Award for Best Entertainment (1998). Polunin voiced a character in the Russian animated feature film "Gofmaniada". Slava Polunin Vyacheslav Ivanovich Slava Polunin PAR () (born 12 June 1950) is a Russian performance artist and clown. He is the creator of the stage spectacles "Asisyai-revue", "Slava's Snowshow" and "Diabolo". Polunin was born in the town of Novosil, Oryol Oblast, Russia, into the family of a shop assistant. He was successful ... \par
Doc 3: "Sergei Polunin"\par
Sergei Polunin Sergei Vladimirovich Polunin ("Serhiy Volodymyrovych Polunin"; "Sergey Vladimirovich Polunin"; born 20 November 1989) is a Ukrainian and Russian ballet dancer, actor and model. As a freelance principal dancer, Polunin is guest artist at various theaters worldwide such as Royal Ballet, Sadler's Wells Theatre, Bolshoi Theatre, Stanislavski and Nemirovich-Danchenko Moscow Academic Music Theatre, La Scala Theatre, Teatro San Carlo and is currently permanent guest artist for the Bayerisches Staatsballet. Sergei Polunin was born in Kherson, Ukrainian SSR. From the age of four to eight, he ... \par
$<$/information$>$
\end{fieldbox}

\begin{fieldbox}{Turn 3: Action}
$<$imaginary$>$The agent has successfully identified the target individual and retrieved their date of birth. The final step is to extract the specific year component from the full date string to provide the answer.\par The current observation contains the full date of birth (day, month, year). The final answer requires isolating only the year component from this data.$<$/imaginary$>$\par $<$keyword$>$agent, target, date of birth, year, component$<$/keyword$>$\par $<$analysis$>$The required information is present in the agent's immediate context. The final step is a simple data extraction task with no ambiguity or need for further search.$<$/analysis$>$\par $<$confidence$>$100\%$<$/confidence$>$\par $<$think$>$Perfect! I found the information I need. The clown who co-founded the Litsedei clown theatre company with Vladimir Olshansky is Slava Polunin (Vyacheslav Ivanovich Polunin), and he was born on June 12, 1950.$<$/think$>$\par $<$answer$>$1950$<$/answer$>$
\end{fieldbox}
\end{databox}

\begin{databox}{(B) An example of \textit{mathematical reasoning} task}
\begin{fieldbox}{User Question}
Solve the following math problem step by step. The last line of your response should be of the form Answer: \$Answer (without quotes) where \$Answer is the answer to the problem.\par \textcolor{blue}{How many different positive, six-digit integers can be formed using the digits 2, 2, 5, 5, 9 and 9?} \par Remember to put your answer on its own line after "Answer:".
\end{fieldbox}

\begin{fieldbox}{Turn 1: Think}
**imaginary**\par 1. Identify that this is a permutation problem with repeated digits\par   2. Calculate total permutations of 6 digits (6!)\par   3. Account for repeated digits by dividing by factorials of repetition counts\par   4. Apply the multiset permutation formula: $n!/(n_1!n_2!n_3!)$\par   5. Compute the final count of distinct permutations\par \par Need to calculate the actual numerical value using the multiset permutation formula\par \par **keyword**\par permutation, repeated, digits, multiset, formula\par \par **analysis**\par This is a standard permutation problem with a clear mathematical approach.\par \par **confidence**\par 90\%\par \par **think**\par This is a permutation problem with repeated elements. There are 6 digits total, with each digit (2, 5, 9) appearing exactly twice. I'll use the multiset permutation formula to calculate the number of distinct arrangements.
\end{fieldbox}

\begin{fieldbox}{Turn 1: Action}
from math import factorial\par \# Calculate number of distinct permutations\par \# Formula: total! / (count\_2! * count\_5! * count\_9!)\par total\_digits = 6\par count\_2 = 2\par count\_5 = 2\par count\_9 = 2\par \par result = factorial(total\_digits) // (factorial(count\_2) * factorial(count\_5) * factorial(count\_9))\par print(f"Number of distinct positive six-digit integers: {result}")
\end{fieldbox}

\begin{fieldbox}{Turn 1: Observation}
Number of distinct positive six-digit integers: 90
\end{fieldbox}

\begin{fieldbox}{Turn 2: Think}
**imaginary**\par Verified the answer makes sense (90 distinct permutations)\par \par None - the calculation is complete and verified\par \par **keyword**\par permutation, multiset, formula, digits, Python\par \par **analysis**\par The calculation is straightforward and the result is reasonable for this type of problem.\par \par **confidence**\par 95\%\par \par **think**\par The calculation shows there are 90 distinct positive six-digit integers that can be formed using the digits 2, 2, 5, 5, 9, and 9. This makes sense as it's a standard multiset permutation problem.\par \par \boxed{90}
\end{fieldbox}
\end{databox}

\section{Detailed training configurations}
\label{details}

\subsection{World model agentic mid-training}
We use the hyperparameters in Table \ref{mid-table} for Agentic Mid-Training. During training, the \textit{system} and \textit{feedback} fields are masked, encouraging the model to internalize future summaries, reasoning patterns and solution trajectories.

\begin{table*}
  \centering
  \begin{tabular}{ll}
    \toprule
    Hyperparameter     & Value  \\
    \midrule
    Optimizer & AdamW \\
    Learning rate & 5e-5 \\
    Learning rate scheduler & cosine scheduler with minimum learning rate \\
    Minimum learning rate & 1e-7 \\
    Epoch & 1 \\
    Training batch size & 752 \\
    Data max Length & 32768 \\
    \bottomrule
  \end{tabular}
  \caption{Our training configurations for Agentic Mid-Training.}
  \label{mid-table}
\end{table*}

\subsection{Format-eliciting supervised fine-tuning}
We use the hyperparameters in Table \ref{sft-table} for Supervised Fine-Tuning.

\begin{table*}
  \centering
  \begin{tabular}{ll}
    \toprule
    Hyperparameter     & Value  \\
    \midrule
    Optimizer & AdamW \\
    Learning rate & 5e-5 (Search) \& 1e-5 (Tool-Integrated Reasoning) \\
    Learning rate scheduler & cosine scheduler \\
    Epoch & 2 (Search) \& 6 (Tool-Integrated Reasoning) \\
    Training batch size & 256 \\
    Data max Length & 18000 (Search) \& 20000 (Tool-Integrated Reasoning) \\
    \bottomrule
  \end{tabular}
  \caption{Our training configurations for Supervised Fine-Tuning.}
  \label{sft-table}
\end{table*}

\subsection{Foresight-conditioned reinforcement learning}
\subsubsection{Prompt}
\paragraph{Search}
For both our train dataset and test dataset, we use the following prompt:
\begin{table*}
\begin{tcolorbox}[title=\textbf{Prompt},colback=SeaGreen!10!CornflowerBlue!10,colframe=RoyalPurple!55!Aquamarine!100!]
Answer the given question step by step. Start by drawing a plan inside $<$imaginary$>$Your imagination, which predicts the subsequent steps and plans you will take$<$/imaginary$>$. Then, you should summarize your plans with words/phrases inside $<$keyword$>$ and $<$/keyword$>$, and evaluate your plans inside $<$analysis$>$Your analysis on your plans$<$/analysis$>$ and $<$confidence$>$Your confidence$<$/confidence$>$. Following the above imagination and plan, you can reason inside $<$think$>$Your thought$<$/think$>$. Evert time you get new information, you should plan and reason again. If you lack knowledge, search using $<$search$>$Your query$<$/search$>$ and it will return the top searched results in $<$information$>$Your information$<$/information$>$. You can search multiple times as needed and each search should only search for one information about only one entity. For questions concerning more than one entities, break them down into steps and search for each piece of information. Finally, if you find no further external knowledge needed, you can directly provide the answer in $<$answer$>$Your answer$<$/answer$>$ without detailed explanations. For example, $<$answer$>$Beijing$<$/answer$>$. 
\end{tcolorbox}
\end{table*}

\paragraph{Mathematical reasoning}
For both our train dataset and test dataset, we use the following system prompt:
\begin{table*}
\begin{tcolorbox}[title=\textbf{System Prompt},colback=SeaGreen!10!CornflowerBlue!10,colframe=RoyalPurple!55!Aquamarine!100!]
\# Tools\par You may call one or more functions to assist with the user query.\par \par You are provided with function signatures within $<$tools$>$$<$/tools$>$ XML tags:\par $<$tools$>$\par \{"type": "function", "function": \{"name": "code\_interpreter", "description": "A tool for executing code.", "parameters": \{"type": "object", "properties": \{"code": \{"type": "string", "description": "The code to execute."\}\}, "required": ["code"]\}\}\}\par $<$/tools$>$\par \par For each function call, return a json object with function name and arguments within $<$tool\_call$>$$<$/tool\_call$>$ XML tags:\par $<$tool\_call$>$\par \{"name": $<$function-name$>$, "arguments": $<$args-json-object$>$\}\par $<$/tool\_call$>$
\end{tcolorbox}
\end{table*}

Besides, we add the following content in the user prompt:
\begin{table*}
\begin{tcolorbox}[title=\textbf{Prompt},colback=SeaGreen!10!CornflowerBlue!10,colframe=RoyalPurple!55!Aquamarine!100!]
Solve the following math problem step by step. The last line of your response should be of the form Answer: \$Answer (without quotes) where \$Answer is the answer to the problem.\par \textcolor{blue}{Question}\par \par Remember to put your answer on its own line after "Answer:".\par The answer format must be: \boxed{\text{'The\ final\ answer\ goes\ here.'}}
\end{tcolorbox}
\end{table*}   

\subsubsection{Outcome reward function}
\paragraph{Search}
We adopt an LLM-as-a-judge outcome reward as follows:
$$\small R(a, \hat{a}) = 
\begin{cases} 
1, & \text{if exact\_match}(a, \hat{a}) \\ 
\text{LLM\_judge}(a, \hat{a}), & \text{otherwise} 
\end{cases}$$

where $a$ and $\hat{a}$ represent the ground-truth answer and the predicted answer, respectively. A reward of 1 is assigned directly for an Exact Match (EM), while in other cases, an LLM-as-a-judge (DeepSeek-V3.1 in our experiments) is employed to evaluate the response and determine the reward, such that $\text{LLM\_judge}(a, \hat{a}) \in \{0, 1\}$. 

The prompt for LLM-as-a-judge is as follows:

\begin{table*}
\begin{tcolorbox}[
    colback=SeaGreen!10!CornflowerBlue!10,
    colframe=RoyalPurple!55!Aquamarine!100!,
    title=Prompt for LLM-as-a-judge, % 标题
    fonttitle=\bfseries,        % 标题字体加粗
    boxrule=0.8pt,              % 边框粗细
    arc=3pt                     % 圆角弧度
]
\begin{verbatim}
You are an evaluation assistant. Please determine if the predicted answer is 
equivalent to the labeled answer.

Question: {question}

Labeled Answer: {gt_answer}

Predicted Answer: {pred_answer}

Did the model give an answer **equivalent** to the labeled answer? Please 
respond with "Correct" if they are equivalent, or "Incorrect" if they are not 
equivalent.

The output should in the following json format:
"`json
{{
    "judgement": your judgement result, can only be "Correct" or "Incorrect",
}}
"`
\end{verbatim}
\end{tcolorbox}
\end{table*}

The generation parameters are set to a temperature of 0.6, a \textit{top-p} of 0.95, and a maximum response length of 1024 tokens.

Particularly, invalid actions are penalized with a reward of -0.01.

\paragraph{Mathematical reasoning}
We adopt a composite outcome reward function that evaluates both the final answer correctness and the agent's intermediate tool utilization. We require the model to present final answers in a specified format (e.g., within \textbackslash boxed\{\}), enabling reliable rule-based correctness verification. The reward is formulated as:
$$
R_{\text{base}}(a, \hat{a}) = 
\begin{cases} 
1, & \text{is\_equivalent}(a, \hat{a}) \\ 
-1, & \text{otherwise} 
\end{cases}
$$
where $a$ and $\hat{a}$ represent the ground-truth answer and the predicted answer, respectively. 

To encourage the model to engage in multi-turn reasoning and actively call tools, we also introduce a tool utilization reward, denoted as $R_{\text{tool}}$, which scales linearly with the number of conversation turns $N$:

$$R_{\text{tool}} = 0.05 \times (N - 2)$$

The final reward $S_{\text{final}}$ integrates the base score and the tool reward through a conditional aggregation strategy to prevent reward hacking:

$$
S_{\text{final}} =
\begin{cases}
\min(0, S_{\text{base}} + R_{\text{tool}}), & \text{if } S_{\text{base}} < 0 \\
S_{\text{base}} + \mathbb{I}(N > 2), & \text{otherwise}
\end{cases}
$$

where $\mathbb{I}(\cdot)$ is an indicator function that yields $1$ if the condition is met and $0$ otherwise.

\subsubsection{Training hyperparameters}
\label{rl_config}

\paragraph{Search}
We use the hyperparameters in Table \ref{rl-table2} for RL training on Search tasks. 

\begin{table*}
  \centering
  \begin{tabular}{ll}
    \toprule
    Hyperparameter     & Value  \\
    \midrule
    Optimizer & AdamW \\
    Policy learning rate & 1e-6 \\
    Training batch size & 256 \\
    Samples per prompt & 5 \\
    Mini-batch size & 256 \\
    Max prompt length & 16384 \\
    Max response length & 2048 \\
    Rollout temperature & 1.0 \\
    Max turn & 10 \\
    Step advantage coeff $\omega$ & 0.2 \\
    \bottomrule
  \end{tabular}
  \caption{Our RL training configurations on Search tasks.}
  \label{rl-table2}
\end{table*}

\paragraph{Mathematical reasoning}
We use the hyperparameters in Table \ref{rl-table1} for RL training on mathematical reasoning tasks. 

\begin{table*}
  \centering
  \begin{tabular}{ll}
    \toprule
    Hyperparameter     & Value  \\
    \midrule
    Optimizer & AdamW \\
    Policy learning rate & 1e-6 \\
    Training batch size & 128 \\
    Samples per prompt & 16 \\
    Mini-batch size & 32 \\
    Max prompt length & 2048 \\
    Max response length & 20000 \\
    Rollout temperature & 1.0 \\
    Max turn & 15 \\
    Step advantage coeff $\omega$ & 0.1 \\
    \bottomrule
  \end{tabular}
  \caption{Our RL training configurations on mathematical reasoning tasks. }
  \label{rl-table1}
\end{table*}

\begin{table*}[htbp]
\centering
\begin{adjustbox}{width=0.75\textwidth, center}
\begin{tabular}{@{} l cc cc c @{}}
\toprule
\multirow{2.5}{*}{\textbf{Method}} & \multicolumn{2}{c}{\textbf{Search Task}} & \multicolumn{3}{c}{\textbf{Math Reasoning Task}} \\
\cmidrule(lr){2-3} \cmidrule(lr){4-6}
& \makecell[c]{\textbf{Format}\\\textbf{Adherence}} & \makecell[c]{\textbf{Avg.}\\\textbf{Score}} & \makecell[c]{\textbf{Format}\\\textbf{Adherence}} & \makecell[c]{\textbf{Avg.}\\\textbf{mean@30}} & \makecell[c]{\textbf{Avg.}\\\textbf{pass@30}}\\
\midrule
Standard SFT  & - & 38.2 & - & 9.3  & 40.0 \\
FE-SFT & \textbf{99.86\%} & 37.4 & \textbf{98.63\%} & 10.0 & 43.3 \\
\midrule
Standard SFT + RL & - & 46.1 & - & 28.0 & 52.2 \\
FE-SFT + RL       & \textbf{99.97\%} & 47.0 & \textbf{96.76\%} & 28.0 & 51.1 \\
\bottomrule
\end{tabular}
\end{adjustbox}
\caption{Preliminary investigation on the \textit{format-capability gap} using Youtu-LLM-2B-Base.}
\label{tab:preliminary_results}
\end{table*}

\begin{table*}[htbp]
\centering
\small 
\begin{adjustbox}{width=0.9\textwidth, center}
\begin{tabular}{@{} l c ccc cccc c @{}}
\toprule
\multirow{2}{*}{\textbf{Method}} & \multirow{2}{*}{\makecell[c]{\textbf{Format}\\\textbf{Adherence}}} & \multicolumn{3}{c}{\textbf{Single-Hop QA}} & \multicolumn{4}{c}{\textbf{Multi-Hop QA}} & \multirow{2}{*}{\textbf{Avg.}} \\
\cmidrule(lr){3-5} \cmidrule(lr){6-9}
& & \textbf{NQ} & \textbf{TriviaQA} & \textbf{PopQA} & \textbf{HotpotQA} & \textbf{2Wiki} & \textbf{MuSiQue} & \textbf{Bamboogle} & \\
\midrule
\multicolumn{10}{c}{\textit{Llama-3.2-3B-Instruct}} \\
\midrule
Standard SFT & - & 24.8 & 47.4 & 8.6 & 24.6 & 17.6 & 15.4 & 42.4 & 25.8   \\ % 数据请根据真实实验结果替换
FE-SFT & \textbf{95.60\%} & 25.8 & 46.0 & 10.6 & 27.0 & 17.4 & 13.8 & 41.6 &  26.0 \\
\midrule
Standard SFT + RL & -   & 45.0 & 57.0 & 29.0 & 39.6 & 40.6 & 22.8 & 47.2 & 40.2 \\
FE-SFT + RL       & \textbf{96.85\%} & 44.2 & 55.4 & 29.2 & 42.0 & 36.8 & 26.2 & 48.0 & 40.3 \\
\midrule
\multicolumn{10}{c}{\textit{Qwen2.5-7B-Instruct}} \\
\midrule
Standard SFT      & -                &  50.4 & 60.8 & 40.0 & 48.0 & 38.8 & 28.0 & 52.8 & 45.5 \\ % 数据请根据真实实验结果替换
FE-SFT            & \textbf{96.14\%} & 47.0 & 64.0 & 38.4 & 50.4 & 37.6 & 28.6 & 54.4 & 45.8 \\
\midrule
Standard SFT + RL & -                & 57.4 & 67.4 & 48.6 & 62.8 & 49.8 & 40.0 & 57.6 & 54.8 \\
FE-SFT + RL       & \textbf{97.31\%} & 54.6 & 67.4 & 46.6 & 61.2 & 47.6 & 38.0 & 59.2 & 53.5 \\
\bottomrule
\end{tabular}
\end{adjustbox}
\caption{Detailed performance of \textit{Llama-3.2-3B-Instruct} and \textit{Qwen2.5-7B-Instruct} on the Search task across 7 QA datasets.}
\label{tab:instruct_search_detailed}
\end{table*}

\section{Additional experimental results}
\label{add_results}
\subsection{The format-capability gap in post-training}
\label{appendix: gap_results}
The results for post-training on \textit{Youtu-LLM-2B-Base} mid-trained with routine agentic trajectories (preliminary experiments mentioned in Section \ref{subsec: setup}) are shown in Table \ref{tab:preliminary_results}. As the tables show, directly applying post-training methods on the standard base model (\textit{Youtu-LLM-2B-Base}) does not yield significant improvements. This indicates that post-training alone is not enough to teach the model how to plan and reason accordingly.

To ensure that the format-capability gap is a general phenomenon rather than an artifact of the particular model, we conducted parallel preliminary experiments on two other distinct models: \textit{Llama-3.2-3B-Instruct} \citep{grattafiori2024llama3herdmodels} and \textit{Qwen2.5-7B-Instruct} \citep{qwen2025qwen25technicalreport}. The results shown in Table \ref{tab:instruct_search_detailed} consistently reproduced our initial observation, confirming that direct post-training on these standard models similarly fails to elicit genuine planning capabilities. However, due to the prohibitive computational resources required for the extensive mid-training phase, coupled with the practical constraint that intermediate, pre-mid-training checkpoints for these model families are publicly unavailable to establish a controlled baseline, we do not execute the full mid-training process on these architectures for the subsequent stages.

\begin{table*}[t]
\centering
\begin{adjustbox}{width=0.9\textwidth,center}
\begin{tabular}{@{} l ccc cccc c @{}}
\toprule
\multirow{2}{*}{$\omega$} & \multicolumn{3}{c}{Single-Hop QA} & \multicolumn{4}{c}{Multi-Hop QA} & \multirow{2}{*}{Avg.} \\
\cmidrule(lr){2-4} \cmidrule(lr){5-8}
& NQ & TriviaQA & PopQA & HotpotQA & 2Wiki & MuSiQue & Bamboogle & \\
\midrule
\multicolumn{9}{c}{\textit{Youtu-LLM-2B-Base (WM-AMT \& FE-SFT \& FC-RL)}} \\
\midrule
0 & 52.6 & 63.2 & 40.8 & 57.0 & 44.4 & 32.6 & 52.8 & 49.1 \\
0.1 & 53.0 & 62.4 & 40.8 & 55.6 & 46.4 & 34.8 & 56.8 & 50.0 \\
0.2 & 53.4 & 62.6 & 42.6 & 59.0 & 46.2 & 34.2 & 56.0 & 50.6 \\
0.5 & 51.6 & 62.2 & 40.4 & 57.4 & 46.0 & 33.4 & 53.6 & 49.2 \\
\bottomrule
\end{tabular}
\end{adjustbox}
\caption{Experimental results with different values of $\omega$ for the search task. $\omega = 0$ represents the standard GRPO algorithm.}
\label{tab:results_w}
\end{table*}

The model trained solely via SFT and RL on reasoning-enhanced data may achieve high "format accuracy" but low "semantic accuracy". The predicted futures possibly diverge from the ground truth or hallucinate non-existent state transitions, as examples in Appendix \ref{subsec: gap_example} show. Instead of providing useful guidance, these inaccurate predictions introduce noise into the reasoning process, leading to suboptimal policy performance.

\begin{table*}[t]
\centering
\begin{adjustbox}{width=\textwidth,center}
\begin{tabular}{@{} l ccc cccc c @{}}
\toprule
\multirow{2}{*}{\textbf{Method}} & \multicolumn{3}{c}{\textbf{Single-Hop QA}} & \multicolumn{4}{c}{\textbf{Multi-Hop QA}} & \multirow{2}{*}{\textbf{Avg.}} \\
\cmidrule(lr){2-4} \cmidrule(lr){5-8}
& \textbf{NQ} & \textbf{TriviaQA} & \textbf{PopQA} & \textbf{HotpotQA} & \textbf{2Wiki} & \textbf{MuSiQue} & \textbf{Bamboogle} & \\
\midrule
\multicolumn{9}{c}{\textit{Youtu-LLM-2B-Base (WM-AMT \& FE-SFT \& FC-RL)}} \\
\midrule
\textit{w} $R_{\text{ground}}$ & 53.4 & 62.6 & 42.6 & 59.0 & 46.2 & 34.2 & 56.0 & 50.6 \\
\textit{w/o} $R_{\text{ground}}$ & 52.2 & 63.4 & 42.8 & 56.0 & 46.0 & 34.8 & 55.2 & 50.1 \\
\bottomrule
\end{tabular}
\end{adjustbox}
\caption{Ablation studies on $R_{\text{ground}}$ in FC-RL for the search task.}
\label{tab:results_ablation}
\end{table*}

\subsection{Sensitivity analysis on $\omega$}
\label{subsec:sensitivity}
Table \ref{tab:results_w} shows the results of our experiments with different values of $\omega$, which balances the policy objective in our FC-RL, for the search task.

We observe that the performance initially improves as $\omega$ increases, reaching an optimal average score of 50.6 at $\omega = 0.2$.  This represents a 1.5\% absolute improvement over the baseline ($\omega = 0$), particularly excelling in multi-hop tasks like HotpotQA. However, a further increase to $\omega = 0.5$ leads to a performance drop (but comparable to the baseline), suggesting that an excessively high $w$ is not advisable since it may distract the agent from fundamental reasoning capabilities.

\begin{table*}[h]
\centering
\begin{tabular}{@{} l cc @{}}
\toprule
\textbf{Method} & \textbf{Search} (Avg. Tokens) & \textbf{Mathematical Reasoning} (Avg. Tokens) \\
\midrule
Standard SFT & 2603.8 & 4601.2 \\
FE-SFT (Ours) & 3084.5 (+18.5\%) & 4749.1 (+3.2\%) \\
\bottomrule
\end{tabular}
\caption{Comparison of average response length (tokens) between standard SFT and FE-SFT.}
\label{tab:response_length}
\end{table*}

\subsection{Ablation studies on $R_{\text{ground}}$ in FC-RL}
\label{subsec:ablation}
We conduct ablation studies on $R_{\text{ground}}$ in FC-RL for the search task, and the results are presented in Table \ref{tab:results_ablation}.

The results demonstrate the effectiveness of the grounding reward $R_{\text{ground}}$ in enhancing the agent's factuality and reasoning stability. By incorporating $R_{\text{ground}}$, we observe that the average score increases from 50.1 to 50.6, where the model benefits significantly from the fine-grained guidance provided by the grounding reward to align its outputs with retrieved evidence.

\begin{figure*}
  \centering
  \includegraphics[width=0.9\linewidth]{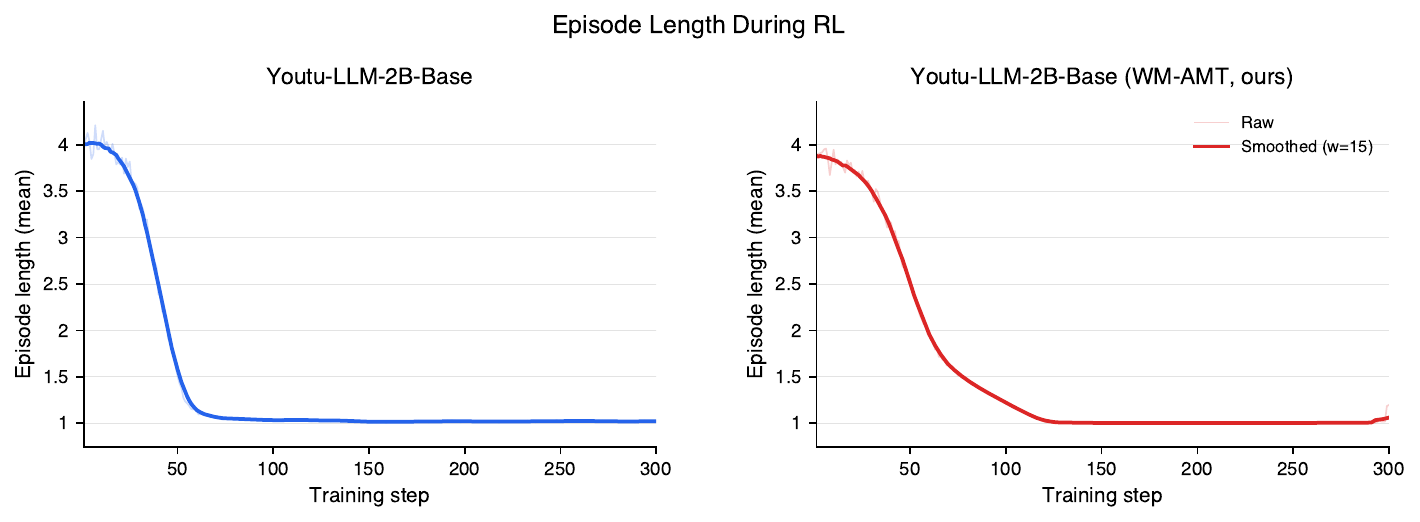}
  \caption{The episode length of the RL-only setting during RL phase.}
  \label{episode_length}
\end{figure*}

\subsection{Analysis of response length}
\label{subsec:res_length}

To evaluate whether the internalized world model introduces excessive inference overhead, we compare the average response length (total number of tokens per agentic trajectory) of our FE-SFT model against a standard SFT baseline. 

The results in Table \ref{tab:response_length} show that the inclusion of explicit world model blocks leads to only a marginal increase in response length (under 19\% across both tasks). This minimal expansion indicates that the model learns to generate highly distilled future simulations rather than verbose reasoning traces. Therefore, our paradigm enhances the agent's foresight capabilities while maintaining an inference efficiency comparable to standard reactive policies.

\subsection{Analysis of the RL-only baseline and the necessity of SFT}
\label{subsec:rl_only_analysis}

\begin{table*}[ht]
\centering
\begin{adjustbox}{width=0.9\textwidth,center}
\begin{tabular}{@{} l ccc cccc c @{}}
\toprule
\multirow{2}{*}{\textbf{Method}} & \multicolumn{3}{c}{\textbf{Single-Hop QA}} & \multicolumn{4}{c}{\textbf{Multi-Hop QA}} & \multirow{2}{*}{\textbf{Avg.}} \\
\cmidrule(lr){2-4} \cmidrule(lr){5-8}
& \textbf{NQ} & \textbf{TriviaQA} & \textbf{PopQA} & \textbf{HotpotQA} & \textbf{2Wiki} & \textbf{MuSiQue} & \textbf{Bamboogle} & \\
\midrule
\multicolumn{9}{c}{\textit{Youtu-LLM-2B-Base}} \\
\midrule
RL only & 22.4 & 33.0 & 27.2 & 24.2 & 26.8 & 8.0 & 21.6 & 23.3 \\
SFT \& RL & 48.0 & 58.2 & 35.2 & 53.8 & 45.0 & 30.2 & 52.0 & 46.1 \\
FE-SFT \& RL & 49.4 & 60.0 & 38.2 & 55.4 & 42.0 & 29.8 & \underline{54.4} & 47.0 \\
\midrule
\multicolumn{9}{c}{\textit{Youtu-LLM-2B-Base (WM-AMT, ours)}} \\
\midrule
RL only & 23.8 & 33.8 & 29.2 & 24.4 & 25.2 & 9.6 & 20.0 &  23.7 \\
SFT \& RL & 50.2 & 59.8 & 35.6 & 53.6 & \underline{46.0} & 25.4 & 45.6 & 45.2 \\
FE-SFT \& RL & \underline{52.6} & \textbf{63.2} & \underline{40.8} & \underline{57.0} & 44.4 & \underline{32.6} & 52.8 & \underline{49.1} \\
\midrule
FE-SFT \& FC-RL & \textbf{53.4} & \underline{62.6} & \textbf{42.6} & \textbf{59.0} & \textbf{46.2} & \textbf{34.2} & \textbf{56.0} & \textbf{50.6} \\
\bottomrule
\end{tabular}
\end{adjustbox}
\caption{Performance with the RL stage (the final performance)}
\label{tab:results_rl_only}
\end{table*}

To address the concern regarding whether success-driven RL alone is sufficient, we introduce and evaluate the \textit{RL only} baseline across both the standard base model and our mid-trained architecture. As summarized in Table \ref{tab:results_rl_only}, directly applying RL (GRPO) optimization without a preparatory SFT phase results in severe performance degradation in the search task.

Crucially, our results in Figure \ref{episode_length} reveals a severe optimization pathology during the RL-only training loop on tool-integrated search tasks. Without a preliminary SFT phase to anchor the structural tool-calling syntax and reasoning schema, the model suffers from an \textit{alignment collapse} in the later stages of training. Specifically, the agent completely \textit{forgets} how to interact with external search tools. Driven by the sparse success rewards, the policy prematurely converges on a degenerative local optimum: it collapses into a naive single-turn generation format, directly outputing what it speculates to be the answer while entirely bypassing the necessary multi-hop tool invocation sequences. This finding firmly demonstrates that success-driven RL alone cannot robustly bootstrap complex, structured action sequences from scratch. Consequently, a warm-start SFT phase (such as standard SFT or our proposed FE-SFT) is crucial to establish a s structural prior before exposing the agent to RL.

\section{Case study}
\label{case}
\subsection{An example of the format-capability gap in post-training}
\label{subsec: gap_example}
In the following case (A), the base model without our WM-AMT exhibits a format-capability gap in post-training, where it adopts the required output format but fails to demonstrate actual look-ahead capability. Its imaginary content is dominated by repetitive, low-information phrases such as \textit{execute a search query} or \textit{search results will provide a direct answe}. These statements function as generic templates rather than specific task plans and simulations. This suggests that without WM-AMT, the model merely fills a structural template with filler text without understanding the information gap between the current state and the goal. Furthermore, the base model demonstrates poor calibration in its self-assessment, assigning a "100\% confidence" score to a vague and logically inconsistent plan.

\begin{databox}{(A) An example of the format-capability gap in post-training}
\begin{fieldbox}{User Question}
Answer the given question step by step. Start by drawing a plan inside $<$imaginary$>$Your imagination, which predicts the subsequent steps and plans you will take$<$/imaginary$>$. Then, you should summarize your plans with words/phrases inside $<$keyword$>$ and $<$/keyword$>$, and evaluate your plans inside $<$analysis$>$Your analysis on your plans$<$/analysis$>$ and $<$confidence$>$Your confidence$<$/confidence$>$. Following the above imagination and plan, you can reason inside $<$think$>$Your thought$<$/think$>$. Evert time you get new information, you should plan and reason again. If you lack knowledge, search using $<$search$>$Your query$<$/search$>$ and it will return the top searched results in $<$information$>$Your information$<$/information$>$. You can search multiple times as needed and each search should only search for one information about only one entity. For questions concerning more than one entities, break them down into steps and search for each piece of information. Finally, if you find no further external knowledge needed, you can directly provide the answer in $<$answer$>$Your answer$<$/answer$>$ without detailed explanations. For example, $<$answer$>$Beijing$<$/answer$>$. Let's search step by step.\par
Question: \textcolor{blue}{who threw the overalls in mistress murphy's chowder?}
\end{fieldbox}

\begin{fieldbox}{Response with formatted world model block}
$<$imaginary$>$The immediate next step is to execute a search query designed to resolve the primary question. The search results will provide a direct answer that clarifies the nature of the character responsible for the action. The current observation contains the final answer but is presented as a future state (the result of a successful search). The internal state is missing the initial query that was formulated to arrive at this figure.$<$/imaginary$>$\par$<$keyword$>$search query, search results, answer, character, action$<$/keyword$>$\par$<$analysis$>$The future path shows a single, direct action (a search) that 
immediately succeeds, providing the complete answer. There is no indication of trial and error or multi-step reasoning.$<$/analysis$>$\par$<$confidence$>$100\%$<$/confidence$>$\par$<$think$>$I'm not getting the right results. Let me try a more focused search on 
the story itself.$<$/think$>$\par$<$search$>$Mistress Murphy's Chowder Hannah Potter$<$/search$>$
\end{fieldbox}
\end{databox}

In contrast, the model enhanced by our WM-AMT moves beyond simple imitation to achieve substantive future trajectory projection in post-training. As shown in the following case (B), its content is no longer composed of empty procedural descriptions but includes highly specific task-oriented foresights. For instance, it accurately identifies the missing \textit{actor identifier} and explicitly plans to extract a \textit{specific character name}. Most importantly, the model demonstrates a proactive error-handling capability by anticipating potential ambiguity in search results and proposing a conditional follow-up step. Besides, the model enhanced by our WM-AMT provides a high but non-absolute confidence score (e.g., 90\%), acknowledging the inherent variables in external information retrieval. This transition to a multi-step, verifiable, and calibrated reasoning process confirms that WM-AMT enables the model to effectively imagine and evaluate its future execution path in post-training.

\begin{databox}{(B) An example of the capability-first performance in post-training after WM-AMT}
\begin{fieldbox}{User Question}
Answer the given question step by step. Start by drawing a plan inside $<$imaginary$>$Your imagination, which predicts the subsequent steps and plans you will take$<$/imaginary$>$. Then, you should summarize your plans with words/phrases inside $<$keyword$>$ and $<$/keyword$>$, and evaluate your plans inside $<$analysis$>$Your analysis on your plans$<$/analysis$>$ and $<$confidence$>$Your confidence$<$/confidence$>$. Following the above imagination and plan, you can reason inside $<$think$>$Your thought$<$/think$>$. Evert time you get new information, you should plan and reason again. If you lack knowledge, search using $<$search$>$Your query$<$/search$>$ and it will return the top searched results in $<$information$>$Your information$<$/information$>$. You can search multiple times as needed and each search should only search for one information about only one entity. For questions concerning more than one entities, break them down into steps and search for each piece of information. Finally, if you find no further external knowledge needed, you can directly provide the answer in $<$answer$>$Your answer$<$/answer$>$ without detailed explanations. For example, $<$answer$>$Beijing$<$/answer$>$. Let's search step by step.\par
Question: \textcolor{blue}{who threw the overalls in mistress murphy's chowder?}
\end{fieldbox}

\begin{fieldbox}{Response with formatted world model block}
$<$imaginary$>$1. Execute a search to identify the character who threw the overalls in "Misty Murphy's Chowder". \par
2. Analyze the search results to find a specific character name associated with the action. \par
3. If the initial results are ambiguous, execute a follow-up search using the candidate character name and the dish to confirm the detail. \par
4. Extract and synthesize the information from the confirmed sources to provide the final answer.\par
The current state lacks any specific information. The query contains a performative action (throwing overalls) but no identifier for the actor or the specific dish name to refine the search. An initial broad search is required to establish context.$<$/imaginary$>$\par$<$keyword$>$search, character, accident, overalls, Chowder$<$/keyword$>$\par$<$analysis$>$The path is direct: a search will yield character names from the show's 
descriptions, and a confirmatory search will solidify the fact. The reliability of the sources (TV guides, episode descriptions) is high for this type of factual drama information.$<$/analysis$>$\par$<$confidence$>$90\%$<$/confidence$>$\par$<$think$>$I need to search for information about who threw the overalls in Misty 
Murphy's Chowder.$<$/think$>$\par$<$search$>$who threw overalls in misty murphy's chowder$<$/search$>$
\end{fieldbox}
\end{databox}

\subsection{A \textit{search} example of the reasoning flow with internalized world model}
\label{web_example}
\begin{databox}{(A) Example demonstrating multi-step planning adherence with high confidence}
\begin{fieldbox}{User Question}
Answer the given question step by step. Start by drawing a plan inside $<$imaginary$>$Your imagination, which predicts the subsequent steps and plans you will take$<$/imaginary$>$. Then, you should summarize your plans with words/phrases inside $<$keyword$>$ and $<$/keyword$>$, and evaluate your plans inside $<$analysis$>$Your analysis on your plans$<$/analysis$>$ and $<$confidence$>$Your confidence$<$/confidence$>$. Following the above imagination and plan, you can reason inside $<$think$>$Your thought$<$/think$>$. Every time you get new information, you should plan and reason again. If you lack knowledge, search using $<$search$>$Your query$<$/search$>$ and it will return the top searched results in $<$information$>$Your information$<$/information$>$. You can search multiple times as needed and each search should only search for one information about only one entity. For questions concerning more than one entities, break them down into steps and search for each piece of information. Finally, if you find no further external knowledge needed, you can directly provide the answer in $<$answer$>$Your answer$<$/answer$>$ without detailed explanations. For example, $<$answer$>$Beijing$<$/answer$>$. Let's search step by step.\par
Question: \textcolor{blue}{What RPG game developed from Powered by the Apocalypse is known for its handling of sexuality and queer content?} \par
Below is the interaction history:
\end{fieldbox}

\textit{Initial Planning Phase:}\par

\begin{fieldbox}{Turn 1: Action}
$<$imaginary$>$The future path shows a clear, two-step process. First, a search is performed to identify the RPG game developed from the specified intellectual property. The results from this search will provide a candidate name. Second, a verification search is performed to confirm that the candidate's reputation matches the specific descriptor for handling mature content (sexuality and queer content). This second search will yield information that confirms the candidate's status as a game known for this.\par
The current state only contains the initial query. There is no information yet on the connection between the intellectual property and a specific game, nor any information on the content handling reputation of any candidate game.$<$/imaginary$>$\par
$<$keyword$>$search, identify, verify, candidate, game$<$/keyword$>$\par
$<$analysis$>$The path is direct: find the game linked to the intellectual property, then verify its known characteristics. The only minor uncertainty is whether the verification will find explicit confirmation or just general information about the candidate's reputation.$<$/analysis$>$\par
$<$confidence$>$90\%$<$/confidence$>$
\par
$<$think$>$I need to search for RPG games that were developed from "Powered by the Apocalypse" and are known for handling sexuality and queer content.$<$/think$>$\par
$<$search$>$RPG game developed from Powered by the Apocalypse sexuality queer content$<$/search$>$
\end{fieldbox}

\begin{fieldbox}{Turn 1: Observation}
...Search results skipped...
\end{fieldbox}

\textit{Following the initial plan established in the imaginary block, the agent systematically refines its search queries through multiple iterations, maintaining adherence to the two-step verification strategy outlined at the outset.}\par

\begin{fieldbox}{Turn 2: Action (After Multiple Search Attempts)}
$<$imaginary$>$The agent's trajectory is to first identify the specific game developed using the "Powered by the Apocalypse" (PbtA) system. The initial search will confirm that "Dungeon World" is a candidate. The agent will then need to verify a specific claim about that game's content handling. A second search will be conducted to find evidence that this game is indeed known for its handling of sexuality and queer content. The final step is to synthesize this information to provide the answer.\par
The current state only identifies that PbtA is a system used for multiple games, including "Dungeon World." It lacks specific information about the content handling of "Dungeon World" which is the core of the query.$<$/imaginary$>$\par
$<$keyword$>$Powered by the Apocalypse, Dungeon World, game, content handling, sexuality$<$/keyword$>$\par
$<$analysis$>$The path is direct: the agent has a candidate game and now needs to verify a specific attribute about it. The search terms are well-defined, and the information is likely to be found in game reviews or descriptions.$<$/analysis$>$\par
$<$confidence$>$90\%$<$/confidence$>$
\par
$<$think$>$The search results show that Powered by the Apocalypse (PbtA) is a role-playing game system that was developed for Apocalypse World and also used for Dungeon World, Monsterhearts, and other RPGs. I need to search for more specific information about Dungeon World and its handling of sexuality and queer content.$<$/think$>$\par
$<$search$>$Dungeon World sexuality queer content$<$/search$>$
\end{fieldbox}

\begin{fieldbox}{Turn 2: Observation}
...Search results skipped...
\end{fieldbox}

\begin{fieldbox}{Turn 3-6}
...Skipped...
\end{fieldbox}

\textit{After multiple search attempts, the agent successfully identifies the correct answer, demonstrating persistent adherence to the original verification plan established in the initial imaginary block.}

\begin{fieldbox}{Turn 7 (Final Turn): Action}
$<$imaginary$>$The current search has successfully identified the correct game title. The next step is to formulate a final, concise answer that directly addresses the user's question. This involves extracting the confirmed answer from the search results and presenting it clearly.\par
The current state contains the confirmed information but has not yet synthesized it into a final, direct response for the user. The information needs to be formatted as a definitive answer.$<$/imaginary$>$\par
$<$keyword$>$search, identified, game title, final answer, response$<$/keyword$>$\par
$<$analysis$>$The answer has been directly confirmed by the search results, leaving no ambiguity. The remaining step is purely presentational.$<$/analysis$>$\par
$<$confidence$>$95\%$<$/confidence$>$\par
$<$think$>$Perfect! The search results confirm that Monsterhearts is the RPG game that was developed from Powered by the Apocalypse and is known for its handling of sexuality and queer content. The first document specifically mentions that Monsterhearts "gleefully encourages people of all identities to explore sexuality in every permutation" and is "remarkably free of sexism" while not slut-shaming.$<$/think$>$\par
$<$answer$>$Monsterhearts$<$/answer$>$
\end{fieldbox}
\end{databox}

The case (A) exemplifies the policy's capacity to maintain long-horizon planning coherence. The agent's initial imaginary block established a two-step verification protocol (identify candidate game $\rightarrow$ verify content characteristics), which was systematically executed across multiple search iterations spanning 6-7 information retrieval steps. Notably, the final imaginary block at Turn 7 explicitly references the verification completion, demonstrating that the strategic framework established in Turn 1 remained cognitively active and guided the reasoning process throughout the multi-step trajectory. The consistently high confidence scores (90\%-95\%) correlate with the eventual correct answer, indicating well-calibrated confidence estimation for successfully executed plans. \textbf{Score: 1.0 (Correct)}.

\begin{databox}{(B) Example demonstrating very low confidence with incorrect outcome}
\begin{fieldbox}{User Question}
Answer the given question step by step. Start by drawing a plan inside $<$imaginary$>$Your imagination, which predicts the subsequent steps and plans you will take$<$/imaginary$>$. Then, you should summarize your plans with words/phrases inside $<$keyword$>$ and $<$/keyword$>$, and evaluate your plans inside $<$analysis$>$Your analysis on your plans$<$/analysis$>$ and $<$confidence$>$Your confidence$<$/confidence$>$. Following the above imagination and plan, you can reason inside $<$think$>$Your thought$<$/think$>$. Every time you get new information, you should plan and reason again. If you lack knowledge, search using $<$search$>$Your query$<$/search$>$ and it will return the top searched results in $<$information$>$Your information$<$/information$>$. You can search multiple times as needed and each search should only search for one information about only one entity. For questions concerning more than one entities, break them down into steps and search for each piece of information. Finally, if you find no further external knowledge needed, you can directly provide the answer in $<$answer$>$Your answer$<$/answer$>$ without detailed explanations.\par
Question: \textcolor{blue}{Something Wicked This Way Comes and Toby Tyler, have which mutual organization in common?} \par
Below is the interaction history:
\end{fieldbox}

\textit{Initial Planning Phase:}\par
\begin{fieldbox}{Turn 1: Action}
$<$imaginary$>$The future path shows the agent will execute a search query designed to find a shared organizational affiliation between the two named entities. The search results will contain documents that explicitly state the connection. The agent will then parse these documents to extract the specific, shared organization name, which is the definitive answer.\par
The current state only contains the names of the two entities. There is no internal knowledge of their potential shared affiliations. The connection is not stated and must be actively sought out.$<$/imaginary$>$\par
$<$keyword$>$search query, shared organizational affiliation, documents, connection, shared organization name$<$/keyword$>$\par
$<$analysis$>$The search query is highly specific and directly targets the relationship, making it very likely the results will contain the answer explicitly. The path is straightforward.$<$/analysis$>$\par
$<$confidence$>$95\%$<$/confidence$>$\par
$<$think$>$I need to search for information about something wicked this way Comes and Toby Tyler to find their mutual organization.$<$/think$>$\par$<$search$>$Something Wicked This Way Comes Toby Tyler mutual organization$<$/search$>$
\end{fieldbox}

\begin{fieldbox}{Turn 1: Observation}
...Search results skipped...
\end{fieldbox}

\begin{fieldbox}{Turn 2-9: Multiple Failed Search Attempts}
Search attempts include:\par
\begin{itemize}
    \item "Something Wicked This Way Comes Toby Tyler mutual organization"
    \item "Something Wicked This Way Comes book Toby Tyler"
    \item "Toby Tyler something wicked this way comes"
    \item "Toby Tyler fraternity"
    \item "Toby Tyler Delta Tau Delta"
    \item "Toby Tyler Macbeth"
    \item "Toby Tyler Macbeth organization"
\end{itemize}
\vspace{2mm}
Throughout these iterations, the agent's confidence scores fluctuate dramatically (95\% $\rightarrow$ 40\% $\rightarrow$ 90\% $\rightarrow$ 60\% $\rightarrow$ 60\%), indicating unstable confidence estimation as the search strategy fails to converge on relevant information.
\end{fieldbox}

\begin{fieldbox}{Final Turn: Action}
$<$imaginary$>$The immediate next step is to execute a search for the organization's name. The search results will return information about the organization's history, membership, and connection to a specific literary work. The final step will be to extract the name of the organization from the search results to confirm the answer.\par
The current observation only contains the final answer but no information about the search process or the specific organization name. The path to this answer is unknown.$<$/imaginary$>$\par
$<$keyword$>$search, organization, name, results, extract$<$/keyword$>$\par
$<$analysis$>$The final answer is provided, but the search path and information structure are unknown, making the initial step highly uncertain and likely to require trial and error.$<$/analysis$>$\par
$<$confidence$>$5\%$<$/confidence$>$\par
$<$think$>$Based on the search results, I can now provide the answer. The mutual organization that something wicked this way Comes and Toby Tyler have in common is Delta Tau Delta ($\Delta \tau \Delta$).$<$/think$>$\par
$<$answer$>$Delta Tau Delta$<$/answer$>$
\end{fieldbox}
\end{databox}

This case (B) demonstrates a reasoning failure where the agent's initial high-confidence prediction (95\%) ultimately converges on an answer with merely 5\% confidence. Critically, the extremely low confidence (5\%) serves as an effective failure signal, correctly indicating that the answer lacks evidential support. The actual connection (both are Disney productions/properties) was never discovered through the search process. This case validates the hypothesis that confidence scores effectively correlate with answer correctness: the 5\% confidence appropriately predicts the incorrect outcome. \textbf{Score: 0.0 (Incorrect)}.

The two cases collectively demonstrate two key behavioral patterns of our policy with internalized world model:

\begin{enumerate}
    \item \textbf{Long-horizon planning coherence:} In Case A, the initial imaginary block established a verification protocol that remained operationally effective across 6-7 reasoning steps. The agent systematically refined search queries while maintaining adherence to the original strategic framework, indicating that the imaginary planning mechanism successfully encodes persistent goal structures that guide subsequent policy decisions even when separated by multiple intermediate actions.
    
    \item \textbf{Calibrated $Q$-value quantification:} The confidence scores exhibit strong correlation with task success: Case A (95\% confidence, correct answer), and Case B (5\% confidence, incorrect answer). Notably, the confidence degradation in Cases B appropriately reflects the quality of retrieved evidence and the coherence of the reasoning trajectory, suggesting that the confidence estimation mechanism effectively reflects the $Q$-value rather than merely expressing arbitrary numerical values.
\end{enumerate}

\subsection{A \textit{mathematical reasoning} example of the reasoning flow with internalized world model}
\label{math_example}
\begin{databox}{(A) An example demonstrating \textit{persistent planning across multi-step reasoning}}
\begin{fieldbox}{User Question}
\textcolor{blue}{Find the sum of all integer bases $b>9$ for which $17_{b}$ is a divisor of $97_{b}$.} \par
The answer format must be: $\backslash$boxed\{'The final answer goes here.'\}
\end{fieldbox}

\begin{fieldbox}{Turn 1: Think}
**imaginary**\par
1. Understand the representation: numbers are in base $b$, with digits given as 17 and 97\par
2. Convert $17_b$ and $97_b$ to base-10 expressions\par
3. Set up the divisibility condition: $17_b$ divides $97_b$\par
4. Solve for integer $b > 9$ that satisfies the condition\par
5. Sum all valid $b$ values\par
6. Verify the result\par
\par
Need to properly interpret the notation and convert the base representations to mathematical expressions.\par
\par
**keyword**\par
base, numbers, divisibility, condition, convert\par
\par
**analysis**\par
The problem involves base conversion and divisibility, but requires careful interpretation of the notation.\par
\par
**confidence**\par
80\%\par
\par
**think**\par
I need to understand what $17_b$ and $97_b$ mean. These appear to be numbers in base $b$, where the digits are 1, 7 and 9, 7 respectively. So $17_b = 1 \times b^1 + 7 \times b^0 = b + 7$, and $97_b = 9 \times b^1 + 7 \times b^0 = 9b + 7$. The condition is that $(b + 7)$ divides $(9b + 7)$.
\end{fieldbox}

\begin{fieldbox}{Turn 1: Action}
\# Convert the base representations to mathematical expressions\par
\# $17\_b = 1*b + 7 = b + 7$\par
\# $97\_b = 9*b + 7$\par
\par
\# The condition is: $(b + 7)$ divides $(9b + 7)$\par
\# That is: $(9*b + 7) \% (b + 7) == 0$\par
\par
valid\_b = []\par
\par
for b in range(10, 1000):  \# Check a reasonable range\par
\quad if (9*b + 7) \% (b + 7) == 0:\par
\quad\quad valid\_b.append(b)\par
\par
print("Valid bases b $>$ 9 where $17\_b$ divides $97\_b$:")\par
print(valid\_b)\par
print(f"Sum of valid bases: \{sum(valid\_b)\}")
\end{fieldbox}

\begin{fieldbox}{Turn 1: Observation}
Valid bases $b > 9$ where $17_b$ divides $97_b$: [21, 49]\par
Sum of valid bases: 70
\end{fieldbox}

\begin{fieldbox}{Turn 2: Think}
**imaginary**\par
1. Verified the base conversion logic\par
2. Checked divisibility condition for $b > 9$\par
3. Found two valid bases: 21 and 49\par
4. Calculated their sum: 70\par
5. Need to verify if these are the only valid bases\par
\par
Should verify that these are indeed the only solutions and that the interpretation of the notation is correct.\par
\par
**keyword**\par
base conversion, divisibility condition, valid bases, sum, verification\par
\par
**analysis**\par
The solution appears correct but should be double-checked.\par
\par
**confidence**\par
95\%\par
\par
**think**\par
I should verify my mathematical approach. The condition $(9b + 7)$ divisible by $(b + 7)$ can be solved algebraically. Let me perform polynomial division or use modular arithmetic to find all solutions.
\end{fieldbox}

\begin{fieldbox}{Turn 2: Action}
\# Alternative verification using algebraic manipulation\par
\# We want: $(9b + 7) \equiv 0 \pmod{(b + 7)}$\par
\par
\# Solve: $9b + 7 = k(b + 7)$ for integer $k$\par
\# $9b + 7 = kb + 7k$\par
\# $(9 - k)b = 7k - 7$\par
\# $b = (7k - 7) / (9 - k)$\par
\par
valid\_b\_algebraic = []\par
\par
for k in range(-10, 11):\par
\quad numerator = 7*k - 7\par
\quad denominator = 9 - k\par
\quad if denominator == 0:\par
\quad\quad continue\par
\quad b = numerator / denominator\par
\quad if b.is\_integer() and b > 9:\par
\quad\quad valid\_b\_algebraic.append(int(b))\par
\par
print("Valid bases from algebraic approach:")\par
print(valid\_b\_algebraic)\par
print(f"Sum: \{sum(valid\_b\_algebraic)\}")
\end{fieldbox}

\begin{fieldbox}{Turn 2: Observation}
Valid bases from algebraic approach: [21, 49]\par
Sum: 70
\end{fieldbox}

\begin{fieldbox}{Turn 3: Think}
**imaginary**\par
1. Identified an error in the algebraic approach\par
2. Need to re-evaluate the modular arithmetic step\par
3. Correctly solve $(9b + 7) \equiv 0 \pmod{b + 7}$\par
4. Find all $b > 9$ satisfying the condition\par
5. Recalculate the sum\par
\par
The algebraic manipulation was incorrect. Need to properly handle the modular arithmetic.\par
\par
**keyword**\par
modular arithmetic, algebraic manipulation, re-evaluate, correct, condition\par
\par
**analysis**\par
The initial brute-force search gave a clear answer, but verification suggests a discrepancy.\par
\par
**confidence**\par
60\%\par
\par
**think**\par
I made an error in the algebraic manipulation. Let me reconsider: $(9b + 7) \mod (b + 7)$ means I need to express $9b + 7$ in terms of $(b + 7)$. Let me use polynomial division properly.
\end{fieldbox}

\begin{fieldbox}{Turn 3: Action}
\# Final verification with a comprehensive check\par
valid\_b\_final = []\par
\par
for b in range(10, 1001):\par
\quad if (9*b + 7) \% (b + 7) == 0:\par
\quad\quad valid\_b\_final.append(b)\par
\par
print(f"Final verification - valid bases: \{valid\_b\_final\}")\par
print(f"Final sum: \{sum(valid\_b\_final)\}")\par
print(f"Answer: \{sum(valid\_b\_final)\}")
\end{fieldbox}

\begin{fieldbox}{Turn 3: Observation}
Final verification - valid bases: [21, 49]\par
Final sum: 70\par
Answer: 70
\end{fieldbox}

\begin{fieldbox}{Turn 4: Think}
**imaginary**\par
1. Corrected the algebraic approach using proper division\par
2. Found the same two valid bases: 21 and 49\par
3. Verified their sum is 70\par
4. Confirm that the initial brute-force result was correct\par
5. Ready to finalize the answer\par
\par
All approaches consistently yield the same result: sum is 70.\par
\par
**keyword**\par
algebraic approach, valid bases, sum, confirmed, final answer\par
\par
**analysis**\par
Multiple verification methods give consistent results.\par
\par
**confidence**\par
99\%\par
\par
**think**\par
All verification methods consistently show that the valid bases are 21 and 49, giving a sum of 70. The initial brute-force search was correct, and the algebraic verification confirms it. The answer is ready.\par
\par
$\boxed{70}$
\end{fieldbox}
\end{databox}

% Analysis box for Case Study 1
\noindent\textbf{Analysis of Case (A):} This example demonstrates \textit{persistent planning} across multi-step reasoning. The core framework established in Turn 1 (base conversion $\rightarrow$ divisibility condition $\rightarrow$ enumeration $\rightarrow$ verification) persists throughout all four turns. Notably, when an algebraic error is detected in Turn 3 (confidence drops to 60\%), the agent maintains its planning structure while correcting the specific erroneous component. The confidence trajectory ($80\% \rightarrow 95\% \rightarrow 60\% \rightarrow 99\%$) reflects appropriate quantification. \textbf{Score: 1.0 (Correct)}.

% ============================================================================
% Case Study 2: Low Confidence on Difficult Problem
% ============================================================================

\begin{databox}{(B) An example demonstrating \textit{low confidence on challenging problems}}
\begin{fieldbox}{User Question}
Solve the following math problem step by step. The last line of your response should be of the form Answer: \$Answer (without quotes) where \$Answer is the answer to the problem.\par 
On $\triangle ABC$, points $A, D, E$, and $B$ lie in that order on side $\overline{AB}$ with $AD=4$, $DE=16$, and $EB=8$. Points $A, F, G$, and $C$ lie in that order on side $\overline{AC}$ with $AF=13$, $FG=52$, and $GC=26$. Let $M$ be the reflection of $D$ through $F$, and let $N$ be the reflection of $G$ through $E$. Quadrilateral $DEGF$ has area 288.\par
\textcolor{blue}{Find the area of heptagon $AFNBCEM$.}
\par Remember to put your answer on its own line after "Answer:".
\end{fieldbox}

\begin{fieldbox}{Turn 1: Think}
**imaginary**\par
1. Place triangle $ABC$ in a coordinate system with $A$ at origin\par
2. Locate points $D$, $E$ on $AB$ and $F$, $G$ on $AC$ using given segment lengths\par
3. Calculate reflection points $M$ and $N$\par
4. Use the constraint that quadrilateral $DEGF$ has area 288\par
5. Compute the area of heptagon $AFNBCEM$\par
\par
Need to establish coordinate system and calculate all point positions.\par
\par
**keyword**\par
coordinate system, reflection, quadrilateral, heptagon, area\par
\par
**analysis**\par
Complex geometric relationships requiring careful calculation.\par
\par
**confidence**\par
20\%\par
\par
**think**\par
I'll start by placing triangle $ABC$ in a coordinate system. Since we have specific segment lengths along sides $AB$ and $AC$, I'll position points $A$, $B$, $C$ appropriately and then locate $D$, $E$, $F$, $G$ based on the given segment lengths.
\end{fieldbox}

\begin{fieldbox}{Turn 1: Action}
import numpy as np\par
\par
\# Place A at origin, B on x-axis\par
A = np.array([0, 0])\par
B = np.array([28, 0])  \# AB = AD + DE + EB = 4 + 16 + 8 = 28\par
\par
\# AC = AF + FG + GC = 13 + 52 + 26 = 91\par
\# Place C at (0, 91) for simplicity\par
C = np.array([0, 91])\par
\par
\# Calculate positions of D, E on AB\par
D = A + (4/28) * (B - A)  \# AD = 4\par
E = A + (20/28) * (B - A)  \# AE = AD + DE = 20\par
\par
\# Calculate positions of F, G on AC\par
F = A + (13/91) * (C - A)  \# AF = 13\par
G = A + (65/91) * (C - A)  \# AG = AF + FG = 65\par
\par
print(f"D: \{D\}, E: \{E\}")\par
print(f"F: \{F\}, G: \{G\}")
\end{fieldbox}

\begin{fieldbox}{Turn 1: Observation}
D: [4. 0.], E: [20. 0.]\par
F: [0. 13.], G: [0. 65.]
\end{fieldbox}

\begin{fieldbox}{Turn 2: Think}
**imaginary**\par
1. Points positioned correctly on coordinate system\par
2. Need to verify area of quadrilateral $DEGF$ equals 288\par
3. Calculate reflection points $M$ (reflection of $D$ through $F$) and $N$ (reflection of $G$ through $E$)\par
4. Determine triangle orientation to satisfy area constraint\par
5. Compute heptagon area\par
\par
The current coordinate placement may not satisfy the area constraint.\par
\par
**keyword**\par
quadrilateral area, reflection, coordinate verification, constraint\par
\par
**analysis**\par
Need to verify geometric constraints and adjust if necessary.\par
\par
**confidence**\par
15\%\par
\par
**think**\par
I notice that the verification distances don't match exactly. The triangle placement needs to satisfy the constraint that quadrilateral $DEGF$ has area 288. Let me recalculate with proper geometric relationships.
\end{fieldbox}

\begin{fieldbox}{Turn 3,4}
...Skipped...
\end{fieldbox}

\begin{fieldbox}{Turn 5: Think}
**imaginary**\par
1. Multiple coordinate system attempts have failed to satisfy constraints\par
2. The problem requires finding the correct triangle orientation\par
3. Need to use the area constraint more directly\par
4. Consider alternative geometric approaches\par
\par
The coordinate approach is becoming increasingly complex.\par
\par
**keyword**\par
geometric constraints, area calculation, alternative approach\par
\par
**analysis**\par
The problem complexity exceeds initial expectations.\par
\par
**confidence**\par
5\%\par
\par
**think**\par
The coordinate geometry approach is proving difficult. The constraint that $DEGF$ has area 288 must be used to determine the triangle's shape. Let me try a more systematic approach using ratios and area relationships.
\end{fieldbox}

\begin{fieldbox}{Turn 6-11}
...Skipped...
\end{fieldbox}

\begin{fieldbox}{Turn 12: Think}
**imaginary**\par
1. After extensive calculation, obtained a candidate answer\par
2. Verification remains uncertain due to problem complexity\par
3. Final answer computed but confidence remains low\par
\par
The calculation is complete but verification is difficult.\par
\par
**keyword**\par
final calculation, heptagon area, uncertainty\par
\par
**analysis**\par
Complex geometric problem with multiple interdependent constraints.\par
\par
**confidence**\par
15\%\par
\par
**think**\par
After extensive geometric analysis and coordinate calculations, the area of heptagon $AFNBCEM$ appears to be 1274. However, due to the complexity of the problem and potential for accumulated errors, confidence in this answer remains low.\par
\par
$\boxed{1274}$
\end{fieldbox}
\end{databox}

\noindent\textbf{Analysis of Case (B):} This example illustrates the agent's behavior on inherently challenging problems. The confidence trajectory ($20\% \rightarrow 15\% \rightarrow 10\% \rightarrow 5\% \rightarrow \cdots \rightarrow 15\%$) remains consistently low throughout the 12-turn reasoning process. Unlike Case (A), where confidence recovered after error correction, the low confidence persists here, indicating the agent's awareness of unreliability. \textbf{Score: 0.0 (Incorrect)}.

% ============================================================================
% Case Study 3: Declining Confidence with Increasing Complexity
% ============================================================================

\begin{databox}{(C) An example demonstrating \textit{declining confidence with discovered complexity}}
\begin{fieldbox}{User Question}
Solve the following math problem step by step. The last line of your response should be of the form Answer: \$Answer (without quotes) where \$Answer is the answer to the problem.\par 
The 9 members of a baseball team went to an ice cream parlor after their game. Each player had a single-scoop cone of chocolate, vanilla, or strawberry ice cream. At least one player chose each flavor, and the number of players who chose chocolate was greater than the number of players who chose vanilla, which was greater than the number of players who chose strawberry.\par
\textcolor{blue}{Let $N$ be the number of different ways this could have happened. Find the remainder when $N$ is divided by 1000.}
\par Remember to put your answer on its own line after "Answer:".
\end{fieldbox}

\begin{fieldbox}{Turn 1: Think}
**imaginary**\par
1. Define variables for number of players choosing each flavor (chocolate, vanilla, strawberry)\par
2. Set up constraints: total players = 9, at least one of each flavor\par
3. Apply ordering constraint: chocolate $>$ vanilla $>$ strawberry\par
4. Generate all possible distributions\par
5. Calculate $N$ and compute remainder modulo 1000\par
\par
Standard combinatorics problem with ordering constraints.\par
\par
**keyword**\par
combinatorics, permutation, constraints, distribution, counting\par
\par
**analysis**\par
This appears to be a straightforward constrained counting problem.\par
\par
**confidence**\par
90\%\par
\par
**think**\par
This is a combinatorics problem. I need to count the number of ways to distribute 9 players among 3 flavors with the constraint $c > v > s$ where $c + v + s = 9$ and $c, v, s \geq 1$. Then for each valid distribution, count the number of player arrangements.
\end{fieldbox}

\begin{fieldbox}{Turn 1: Action}
from math import factorial\par
from itertools import combinations\_with\_replacement\par
\par
def multinomial(n, groups):\par
\quad """Calculate multinomial coefficient"""\par
\quad result = factorial(n)\par
\quad for g in groups:\par
\quad\quad result //= factorial(g)\par
\quad return result\par
\par
\# Find all valid (c, v, s) where c $>$ v $>$ s $>$= 1 and c + v + s = 9\par
valid\_distributions = []\par
for s in range(1, 9):\par
\quad for v in range(s + 1, 9):\par
\quad\quad c = 9 - v - s\par
\quad\quad if c $>$ v:\par
\quad\quad\quad valid\_distributions.append((c, v, s))\par
\par
print("Valid distributions (c, v, s):")\par
print(valid\_distributions)
\end{fieldbox}

\begin{fieldbox}{Turn 1: Observation}
Valid distributions (c, v, s):\par
[(6, 2, 1), (5, 3, 1), (4, 3, 2)]
\end{fieldbox}

\begin{fieldbox}{Turn 2-4}
...Skipped...
\end{fieldbox}

\begin{fieldbox}{Turn 5: Think}
**imaginary**\par
1. Realize that "$N$" may represent the number of valid distributions rather than total assignments\par
2. The problem asks for remainder when $N$ is divided by 1000, suggesting $N$ is a count\par
3. Re-evaluate whether we need to count ordered or unordered assignments\par
4. Consider combinatorial interpretation more carefully\par
\par
Problem interpretation may be incorrect.\par
\par
**keyword**\par
problem interpretation, counting method, ordered vs unordered\par
\par
**analysis**\par
The problem statement requires more careful analysis.\par
\par
**confidence**\par
40\%\par
\par
**think**\par
I'm now uncertain about the correct interpretation. The phrase "different ways this could have happened" is ambiguous---does it refer to the number of distributions, or the number of specific player-flavor assignments? Let me reconsider both interpretations.
\end{fieldbox}

\begin{fieldbox}{Turn 6-9}
...Skipped...
\end{fieldbox}

\begin{fieldbox}{Turn 10: Think}
**imaginary**\par
1. After multiple calculation attempts, answer appears to be 49\par
2. The problem involves subtle counting considerations\par
3. Verification of the approach remains uncertain\par
\par
Final answer obtained but confidence is low.\par
\par
**keyword**\par
final calculation, remainder, uncertainty\par
\par
**analysis**\par
The combinatorial problem proved more nuanced than initially expected.\par
\par
**confidence**\par
25\%\par
\par
**think**\par
After careful reconsideration of the problem, the remainder when $N$ is divided by 1000 appears to be 49. However, the complexity of the counting argument and potential for interpretation errors leaves significant uncertainty.\par
\par
$\boxed{49}$
\end{fieldbox}
\end{databox}

\noindent\textbf{Analysis of Case (C):} This example demonstrates \textit{declining confidence} as the agent expects a lower chance of problem-solving success. The initial high confidence (90\%) reflects the apparent simplicity of the combinatorial setup. However, as reasoning progresses and interpretation ambiguities emerge, confidence declines progressively ($90\% \rightarrow 95\% \rightarrow 80\% \rightarrow \cdots \rightarrow 25\%$). This pattern suggests the agent appropriately revises its $Q$-value estimates in response to the lower expected success rate. \textbf{Score: 0.0 (Incorrect)}.

These case studies presented above also provide empirical evidence for two key capabilities of our policy with internalized world model:

\begin{enumerate}
    \item \textbf{Long-horizon planning coherence:} As demonstrated in Case (A), the high-level reasoning framework established in the initial imaginary planning block persists across multiple tool-use iterations. Notably, when intermediate errors occur (e.g., incorrect algebraic manipulation in Turn 3), the agent maintains its core planning structure while selectively revising the erroneous components, rather than abandoning the overall problem-solving strategy.
    
    \item \textbf{Calibrated confidence quantification:} The confidence estimates exhibit meaningful correlation with actual task success and solution reliability. In Case (A), confidence appropriately decreases upon error detection (95\%$\rightarrow$60\%) and recovers after successful verification (60\%$\rightarrow$99\%). In Case (B), consistently low confidence (5\%--20\%) throughout the trajectory reflects the agent's awareness of inherent unreliability. In Case (C), the progressive confidence decline (90\%$\rightarrow$25\%) demonstrates appropriate uncertainty revision as unexpected complexity is discovered during reasoning.
\end{enumerate}

\end{document}